\documentclass[10pt,twocolumn,letterpaper]{article}
  
\usepackage[pagenumbers]{cvpr} 

\usepackage[table,xcdraw]{xcolor}

\newcommand{\NAME}{Edit-by-Track}

\newcommand{\topic}[1]
{
\vspace{1mm}\noindent\textbf{#1}
}

\usepackage{soul} 
\usepackage{multirow}
\usepackage{amsmath, amssymb}
\usepackage{pifont}
\usepackage{subcaption}
\usepackage{enumitem, array}
\newcolumntype{C}[1]{>{\centering\arraybackslash}p{#1}}
\newcolumntype{L}[1]{>{\raggedright\arraybackslash}p{#1}}
\newcolumntype{R}[1]{>{\raggedleft\arraybackslash}m{#1}}

\usepackage{blindtext}
\usepackage{graphicx}
\usepackage{caption}
\usepackage{adjustbox} 
\usepackage{multirow} 
\newcommand{\mpage}[3]
{
\begin{minipage}[#1]{#2\linewidth}\centering
#3
\end{minipage}
}

\usepackage[symbol]{footmisc}

\usepackage{upgreek}
\def\vid{V}  
\def\trk{\mathcal{T}}  
\def\cam{\mathcal{P}}  
\def\trktok{\uptau}  
\def\vidtok{\nu}  
\def\latent{\mathbf{z}}  
\def\src{\mathrm{src}}  
\def\tgt{\mathrm{tgt}} 
\def\proj{\mathrm{proj}}  
\def\step{t}  

\def\nframe{F}  
\def\imgh{H}    
\def\imgw{W}    
\def\tokf{f}    
\def\tokh{h}    
\def\tokw{w}    
\def\tokd{d}    
\def\tokfhw{\tokf\tokh\tokw}    
\def\ntrack{N}  
\def\pe{\rho}  
\def\zpe{\sigma}  
\def\attn{\mathrm{Attn}}  

\def\sampled{\mathrm{sampled}}
\def\grid{\mathcal{G}}
\def\RR{\mathbb{R}}

\definecolor{bboxred}{HTML}{FF6875}
\definecolor{bboxred}{HTML}{FF6875}
\definecolor{scoreyellow}{HTML}{FFEEBF}
\definecolor{scorered}{HTML}{FFCCC9}

\definecolor{cvprblue}{rgb}{0.21,0.49,0.74}
\usepackage[pagebackref,breaklinks,colorlinks,allcolors=cvprblue]{hyperref}

\title{Generative Video Motion Editing with 3D Point Tracks}

\author{
Yao-Chih Lee$^{1,3,*}$ \hspace{0.9cm}
Zhoutong Zhang$^{2}$ \hspace{0.9cm}
Jiahui Huang$^{1}$ \hspace{0.9cm}
Jui-Hsien Wang$^{1}$ \\
Joon-Young Lee$^{1}$ \hspace{0.9cm}
Jia-Bin Huang$^{3}$ \hspace{0.9cm}
Eli Shechtman$^{1}$ \hspace{0.9cm}
Zhengqi Li$^{1}$
\vspace{0.2cm} \\
$^1$Adobe Research \hspace{0.8cm}
$^2$Adobe \hspace{0.8cm}
$^3$University of Maryland College Park 
\vspace{0.1cm}
\\
\url{https://edit-by-track.github.io}
\vspace{-0.8cm}
}
\begin{document}
\twocolumn[{
\renewcommand\twocolumn[1][]{#1}
\maketitle
\begin{center}
    \centering
    \captionsetup{type=figure}
    \includegraphics[trim={0 0 0 4mm},clip,width=.97\linewidth]{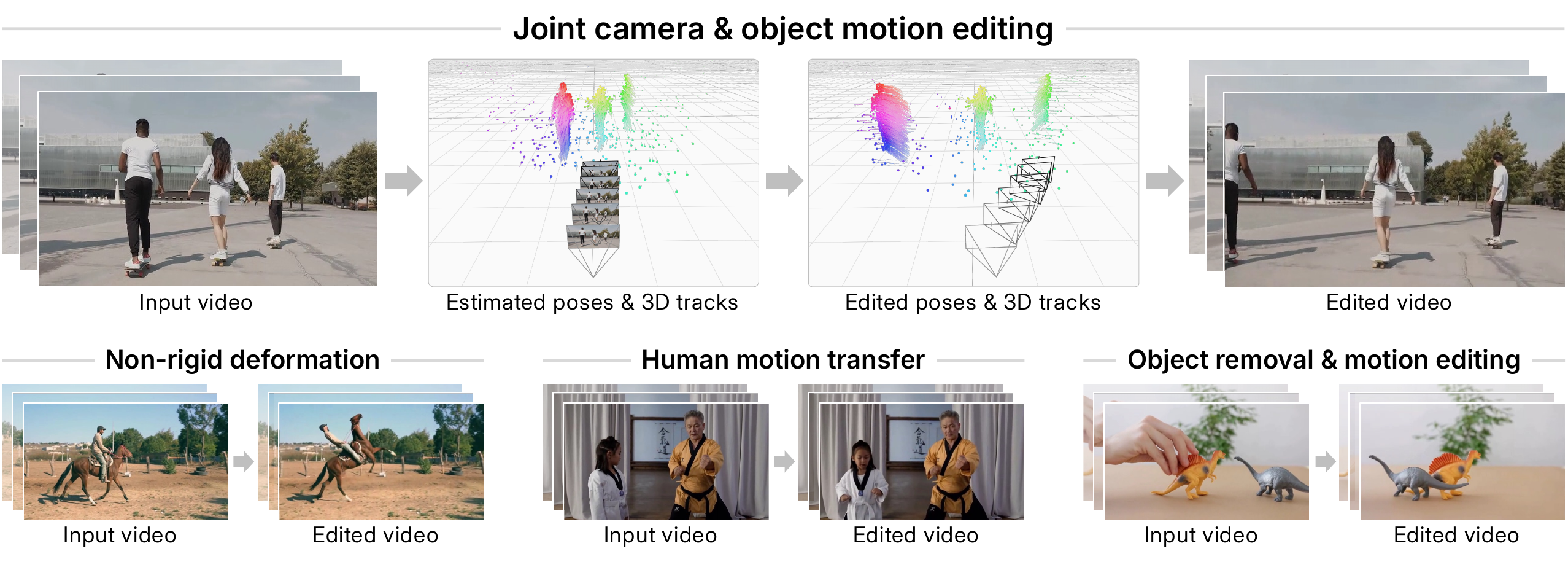}\vspace{-4mm}
    \captionof{figure}{
    \textbf{\NAME{}.}
    Our novel framework enables precise video motion editing via 3D point tracks. By specifying desired 3D trajectories, users can seamlessly control joint camera and object motion, remove objects, and transfer motion between videos. 
    }
    \label{fig:teaser}
    \vspace{-0.5mm}
\end{center}
}]
\begin{abstract}

Camera and object motions are central to a video's narrative.
However, precisely editing these captured motions remains a significant challenge, especially under complex object movements.
Current motion-controlled image-to-video (I2V) approaches often lack full-scene context for consistent video editing, 
while video-to-video (V2V) methods provide viewpoint changes or basic object translation, but offer limited control over fine-grained object motion.
We present a track-conditioned V2V framework that enables joint editing of camera and object motion.
We achieve this by conditioning a video generation model on a source video and paired 3D point tracks representing source and target motions.
These 3D tracks establish sparse correspondences that transfer rich context from the source video to new motions while preserving spatiotemporal coherence.
Crucially, compared to 2D tracks, 3D tracks provide explicit depth cues, allowing the model to resolve depth order and handle occlusions for precise motion editing.
Trained in two stages on synthetic and real data, our model supports diverse motion edits, including joint camera/object manipulation, motion transfer, and non-rigid deformation, unlocking new creative potential in video editing.
\end{abstract}    
\section{Introduction}
\label{sec:intro}

Video motion, a complex blend of scene dynamics and camera movement, conveys rich visual information and supports storytelling.
Nonetheless, precisely editing the motion of a given video--whether correcting camera trajectories or synthesizing different motions for an object--remains challenging, particularly when handling both camera motion and complex object movements with occlusions.

\begin{figure}
\centering
\includegraphics[width=\linewidth]{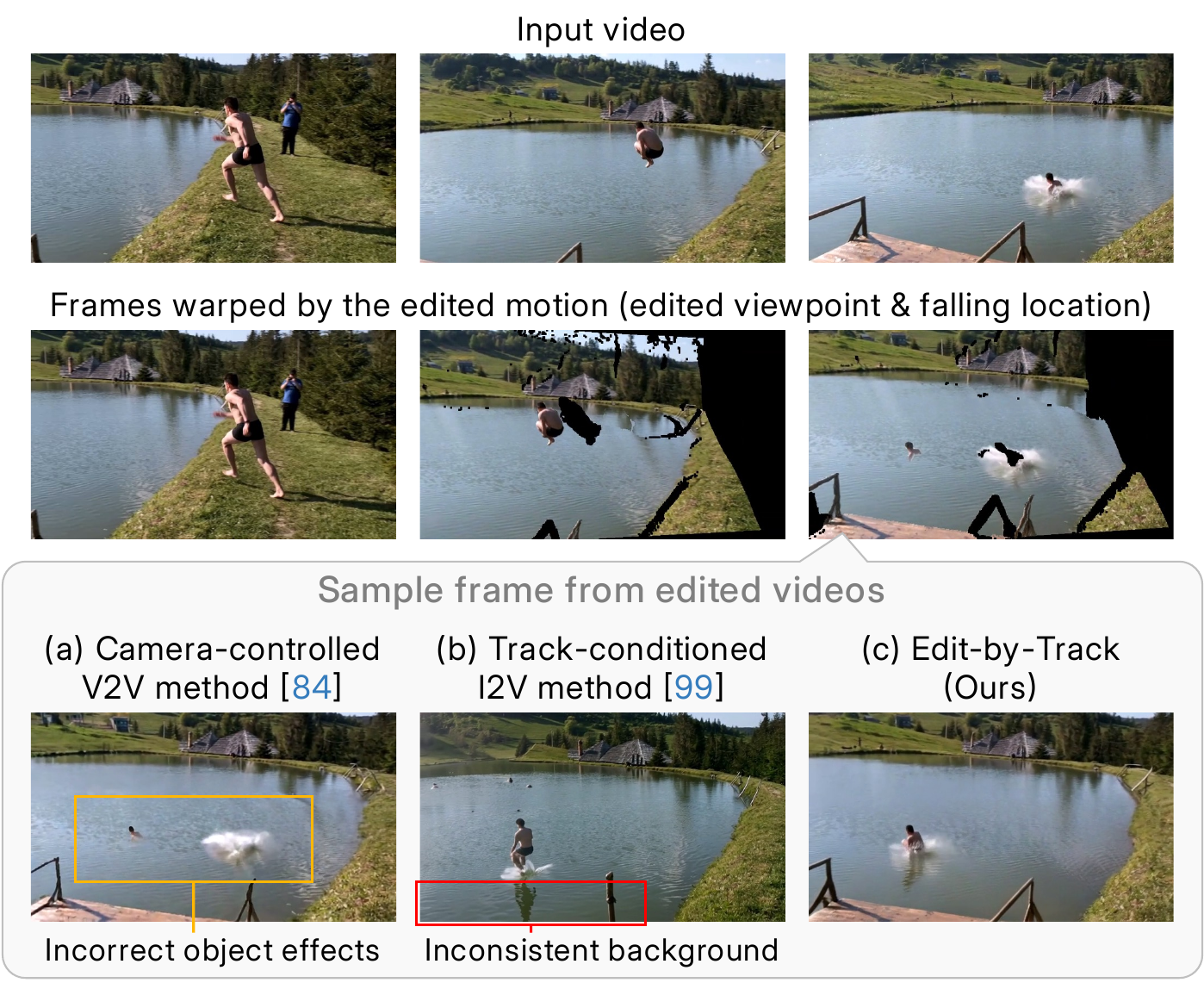}\vspace{-3.5mm}
\caption{\textbf{Limitations of existing methods.}
We demonstrate joint camera and object motion editing on an input video (first row)—changing both the camera viewpoint and the person’s falling location—using frames warped by the edited motion as reference (second row).
The prior camera-controlled V2V approach~\cite{gen3c} inpaints from the warped input video but fails to correct secondary effects (e.g., splashes) caused by the edited object motion.
The track-conditioned I2V method~\cite{ati} loses input scene context by conditioning only on the first frame.
In contrast, our approach edits both camera and object motion while preserving the input context and maintaining coherent causal effects (third row).
}
\vspace{-4mm}
\label{fig:baseline_limitations}
\end{figure}

Existing video editing methods typically address only one side of this problem.
Object-centric approaches~\cite{magicstick,shape-for-motion} enable simple object motion editing (\eg, shifting and resizing) using bounding boxes or 3D meshes, but lack control over camera viewpoints.
Conversely, recent video diffusion models~\cite{trajcrafter,recammaster,gen3c} address camera motion manipulation for videos.
However, these methods are designed to preserve the scene's object motion identically by training on synchronous, multi-view data. 
Hence, they inherently fall short when attempting to edit object motion (Fig.~\ref{fig:baseline_limitations}a).

\footnotetext{*Work done while Yao-Chih was an intern at Adobe Research.}
In this paper, we present, \NAME{}, a unified framework for manipulating \emph{both camera and object motion} in a video.
Inspired by recent track-conditioned image-to-video (I2V) generation models~\cite{motionprompting,diffusion-as-shader},
we adopt point trajectories~\cite{sand2008particle,harley2022particle,tapip3d} as a general motion representation to guide the editing of both types of motions.
Nevertheless, existing track-conditioned I2V models are unsuitable for video editing since they generate motion from \emph{a single frame}, neglect the remaining frames, and thus lose the original scene context (Fig.~\ref{fig:baseline_limitations}b). 
Hence, we propose a motion-controlled video-to-video (V2V) model that leverages the full input video and its corresponding 3D point tracks, establishes sparse correspondences between the source and target motions, and enables precise motion editing while faithfully preserving the original scene context.

Editing 3D motion in a video often exposes previously unseen regions.
To address this challenge, we leverage the strong generative prior of a pretrained text-to-video (T2V) diffusion model~\cite{wan} and further fine-tune it using a new 3D-track conditioner tailored for video editing.
To effectively encode 3D tracks, our 3D-track conditioner uses cross-attention to perform a learnable sampling-and-splatting process that adaptively gathers visual context from the input video and projects it into the target frame space with 3D awareness, while remaining robust to noisy inputs.

Training a V2V diffusion model requires paired videos with specific camera and object motion manipulations, along with 3D point-track annotations—data that is difficult to obtain in real-world settings.
We address this challenge with a two-stage fine-tuning strategy.
The first stage uses synthetic video pairs of animated humans with ground-truth 3D tracks to bootstrap motion control from a pretrained T2V model.
To bridge the domain gap, the second stage further fine-tunes the model on diverse real-world video pairs constructed by sampling non-contiguous clips from monocular videos.
Our model achieves precise, 3D-aware control over both camera and object motion, enabling a wide range of editing capabilities, including manipulating object depth and handling occlusions (Fig.~\ref{fig:teaser}). Extensive experiments demonstrate that our method substantially outperforms state-of-the-art approaches in editing both camera motion and object dynamics, while preserving the coherence of the input context throughout the edited videos.

In summary, our contributions include:
(i) A robust video motion editing model based on a novel 3D track conditioner,
(ii) a new two-stage training pipeline that enables model to learn precise 3D motion control for real video clips,
and (iii) we show our system enables versatile motion editing effects that cannot be done by recent work.

\begin{figure*}
\centering
\includegraphics[clip,trim={2mm 0 0 0},width=.98\linewidth]{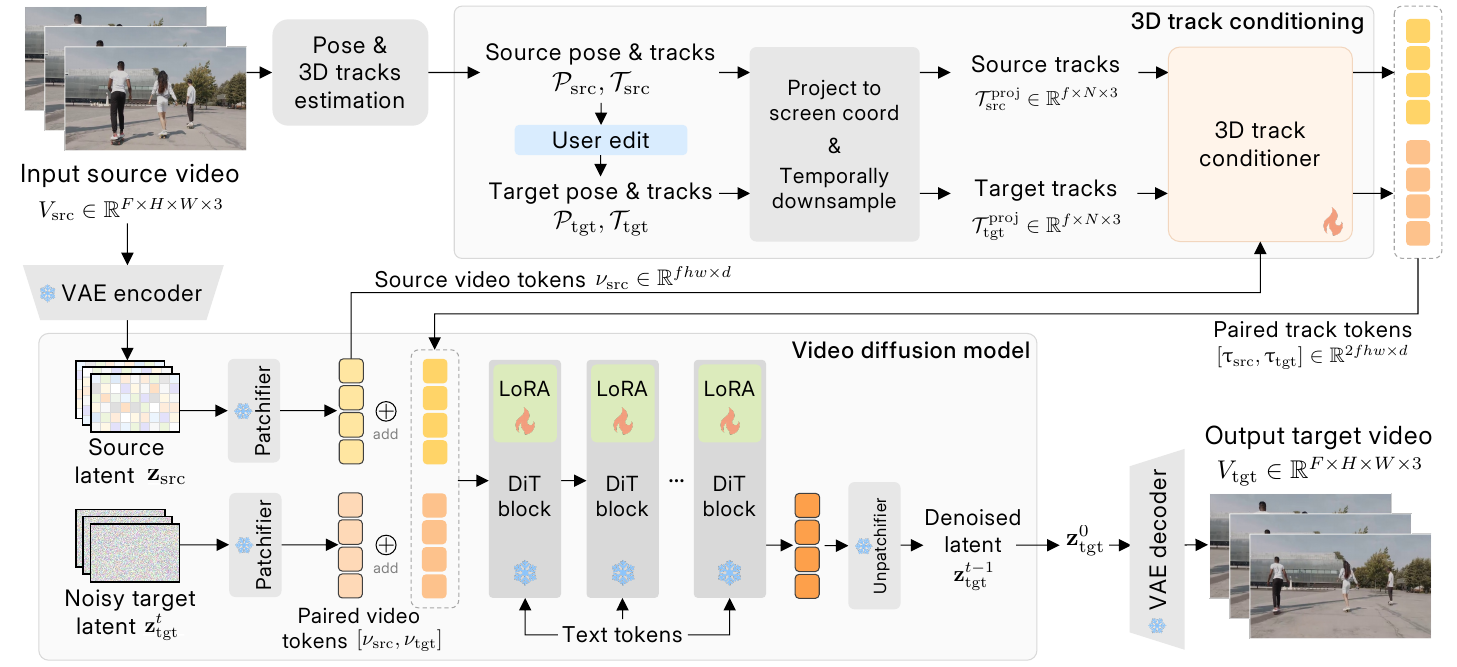}\vspace{-3mm}
\caption{\textbf{\NAME{} framework.}
Given a video $\vid_\src$, we estimate camera poses and 3D tracks using off-the-shelf models~\cite{spatracker2,tapip3d}.
Users then edit the poses and tracks to form pairs of source and target screen-projected tracks, which serve as the motion condition.
To preserve the original visual context, the source video is encoded by a VAE and patchifier into source tokens $\vidtok_\src$, which are concatenated with noisy target tokens $\vidtok_\tgt$.
For motion control, our 3D track conditioner (Fig.~\ref{fig:trackconditioner}) processes the track pairs into tokens $[\trktok_\src, \trktok_\tgt]$, which are added to the paired video tokens $[\vidtok_\src, \vidtok_\tgt]$.
These motion-conditioned tokens are then fed into DiT blocks to denoise $\vidtok_\tgt$.
}
\vspace{-4mm}
\label{fig:framework}
\end{figure*}

\section{Related Work}
\label{sec:related}

\topic{Motion-controlled video generation.}
Recent advances in text-to-video (T2V) and image-to-video (I2V) models~\cite{stablevideodiff,dynamicrafter,videocrafter1,animatediff,cogvideo,cogvideox,wan,hunyuan,lumiere,walt,cosmos,moviegen,snapvideo} have laid the foundation for controlled video synthesis.
A growing body of work~\cite{motionctrl,cameractrl,wang2025actionprompts,draganything,dragnuwa,mofa,cinemaster,motionprompting,perception-as-control,synfmc,diffusion-as-shader,motioncanvas,trajattn,magicmotion,gowiththeflow,frameinnout,motionpro} integrates explicit motion conditions, often through adapters such as ControlNet~\cite{controlnet}, to control the camera and scene dynamics in the generated videos.

For effective motion control, the choice of motion representation is crucial.
For camera motion,~\cite{motionctrl,cameractrl} directly encode and inject camera pose parameters into video models.
For object motion,~\cite{motioncanvas,magicmotion,trailblazer} leverage 2D bounding boxes to guide the object motion.
Alternatively, several approaches~\cite{motionctrl,draganything,dragnuwa,motionprompting,perception-as-control,diffusion-as-shader,motioni2v,trajattn,tora,ati,motioncanvas} adopt point tracks as a unified representation for both camera and object motion, where background point movements correspond to camera motion and foreground points to object motion.
Such point-track conditioning can rely on a few points (\eg, $<10$) for user-friendly input~\cite{motionctrl,perception-as-control,dragnuwa,motioni2v,levitor,tora} or denser, hundreds of points for fine-grained control~\cite{motionprompting,diffusion-as-shader,track4gen}.
Recent studies~\cite{perception-as-control,diffusion-as-shader,levitor} use 3D point tracks for 3D-aware motion synthesis.

However, these I2V models~\cite{motionprompting,diffusion-as-shader,ati,perception-as-control} are not well-suited for video editing tasks--they use \emph{only the first frame} of an input video as a reference, omitting subsequent frames and losing the context of the whole dynamic scene.
In contrast, our V2V framework conditions on the entire input video, preserving the full context of the original scene.

\topic{Camera-controlled video-to-video synthesis.}
Editing camera viewpoints of an input video, often known as novel view synthesis, is a longstanding research topic.
Previous reconstruction-based methods~\cite{nsff,gao2021dynamic,xian2021space,nerfies,dynibar,rodynrf,casualfvs,mosca,shapeofmotion,gaussianmarbles} lack generative priors to hallucinate unseen areas, limiting the effective range of viewpoint changes.

Recent studies leverage generative models to synthesize complete content.
Multi-view diffusion models~\cite{stablevirtualcam,syncammaster,cat4d,4realvideo,dimensionx,lyra} input a video to generate multi-view videos, which can be lifted into a 4D Gaussian Splatting field~\cite{cat4d,4realvideo,dimensionx,lyra}.
Inpainting-based methods~\cite{nvssolver,viewcrafter,reangle,trajcrafter,gen3c,followyourcreation} warp input pixels to the target views, then employ video diffusion models to fill the missing pixels. 
Instead, ReCamMaster~\cite{recammaster} uses a V2V model that directly takes the input video and target camera poses to generate target videos.

However, these methods focus only on camera viewpoint changes, neglecting object motion editing. 
While inpainting approaches may handle a warped video of edited objects, they inherently struggle to synthesize plausible outputs, as the models are trained on synchronous video pairs~\cite{recammaster,kubric4d}. 
In contrast, our method learns from asynchronous video pairs to enable joint camera and object motion editing.

\topic{Video editing.}
Existing studies have made remarkable progress in editing tasks, including object replacement/appearance editing~\cite{videop2p,magicedit,vace,tokenflow,nla,inve,text2live,flowvid,text2videozero,genprop,fatezero,diffusion-motion-transfer,loraedit,i2vedit,dive,anyv2v,rave,ccedit,stablevideo,dynvideoe,vmc,flatten,reenact}, object insertion~\cite{videoanydoor,vace,dynvfx,genprop}, and object removal~\cite{vace,genprop,gen-omnimatte}.
Nevertheless, these methods alter appearance or composition while \emph{preserving the original motion} in output videos. 

Recent works have explored aspects of motion or shape editing in videos, but each addresses only a subset of the broader problem.
Drag-based approaches~\cite{drag-a-video,dragvideo} enable simple shape deformation but struggle with complex motion editing.
Human motion editing methods~\cite{edityourmotionn,motioneditor} condition on human skeleton poses and thus do not generalize to general objects.
Object-centric methods manipulate object position or shape using bounding boxes~\cite{magicstick}, 3D meshes~\cite{shape-for-motion}, or assume static objects~\cite{videohandles}, but lack control over camera or fine-grained object motion.
GS-DiT~\cite{gsdit} and ReVideo~\cite{revideo} use video diffusion models to refine a coarse video draft, generated from a pseudo-Gaussian Field or a masked background with a first-frame edit, respectively.
However, these methods remain sensitive to the rendering quality of the coarse input~\cite{gsdit} and still lose the full scene context of the original video~\cite{revideo}.
In contrast, we introduce a streamlined V2V framework for fine-grained, joint camera and object motion editing.
Our method directly uses the entire unmasked input video and the pair of input and target 3D tracks to generate the desired video.

\topic{Video pose, depth, and 3D track estimation.}
Recent progress in video depth and pose estimation~\cite{vggt,monst3r,casualsam,megasam,pi3,vipe,cut3r} enable 3D point trackers~\cite{delta3dtrack,spatracker2,seurat,tapip3d,uni4d,st4rtrack,spatracker} to lift 2D tracks~\cite{cotracker3,tapir,kim2025exploring,cho2024local,cotracker,omnimotion} into 3D space.
Building on these advances, our method utilizes 3D pose and track estimations to enable user-controllable editing of both camera and 3D object motion.

\section{Method}
\label{sec:method}

Our goal is to generate a target video $\vid_{\tgt} \in \RR^{\nframe\times\imgh\times\imgw\times 3}$ from an input source video $\vid_{\src}$ that accurately reflects user-specified motion editing.
To achieve fine-grained control over both the camera viewpoint and scene dynamics, we leverage a unified motion representation based on 3D point tracks.
We first estimate the source video's per-frame camera parameters $\cam_{\src}$ and $\ntrack$ 3D point tracks $\trk_{\src} \in \RR^{\nframe\times\ntrack\times 3}$ using off-the-shelf video-depth-pose and 3D tracking models~\cite{spatracker2, tapip3d}.
The user then defines the target motion by editing the source camera/object motion $(\cam_{\src}, \trk_{\src})$ to obtain the desired target parameters, $(\cam_{\tgt}, \trk_{\tgt})$.
Subsequently, the source video and the corresponding source and target motion parameters are input to our V2V framework to synthesize the target video (Fig.~\ref{fig:framework}).

Since 3D motion editing inevitably requires synthesizing unseen content, we leverage the strong generative prior of a pretrained T2V generation model~\cite{wan} to fine-tune our V2V model with a 3D track conditioner (Sec.~\ref{sec:framework}).
A major challenge lies in the scarcity of ideal, annotated video pairs for motion manipulation.
To address this, we adopt a two-stage fine-tuning approach (Sec. \ref{sec:training}): We first train the model on synthetic data to establish the core motion control capability, followed by fine-tuning on real-world videos to significantly enhance generalizability.
Finally, we detail how our trained model enables various editing tasks (Sec. \ref{sec:inference}).


\subsection{Track-Conditioned Video-to-Video Model}
\label{sec:framework}
Our \NAME{} is built upon the pretrained T2V model, Wan-2.1~\cite{wan}, a transformer-based video diffusion model (DiT) trained with the Rectified Flow objective~\cite{rectflow} to generate 81-frame videos. 
The T2V generation begins by patchifying a noisy video latent at step $\step$, $\latent_{\tgt}^{\step}$, into video tokens $\vidtok_\tgt \in \RR^{\tokfhw \times \tokd}$. 
These tokens are fed into a series of DiT blocks, conditioned by encoded text tokens, and then unpathchfied back to the denoised latent, $\latent_\tgt^{\step-1}$.
Throughout an iterative denoising process, the final clean latent, $\latent_\tgt^0$ is decoded back to the target RGB video, $\vid_{\tgt}$ by the VAE decoder.
Here, $\tokf$, $\tokh$ and $\tokw$ stand for the downsampled temporal and spatial dimensions, and $\tokd$ is the token dimension.

For video editing, our V2V model is conditioned on a source video, $\vid_\src$, which is encoded by a VAE encoder to a latent, $\latent_\src$, and patchified into source video tokens, $\vidtok_\src$.
These tokens are concatenated with the noisy target video tokens, $\left[\vidtok_\src, \vidtok_\tgt \right]\in\RR^{2\tokfhw\times\tokd}$, to serve as the video condition for the DiT blocks to generate the target video.

\begin{figure}
\centering
\includegraphics[clip,trim={2mm 0 2mm 2mm},width=\linewidth]{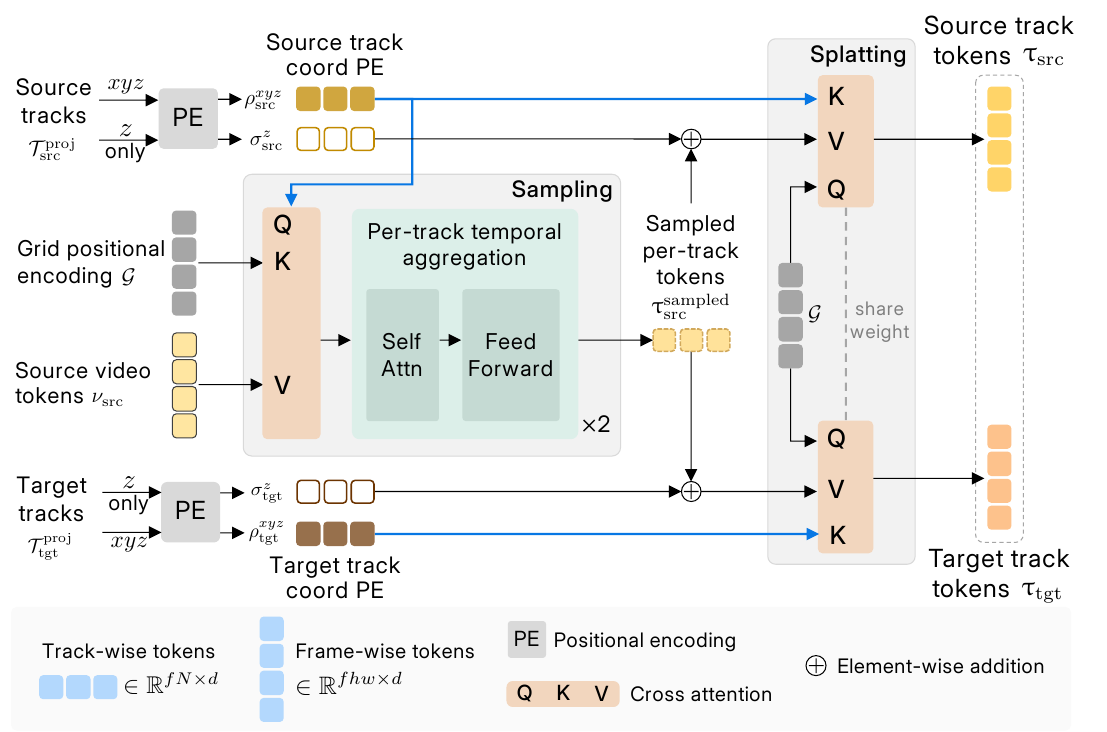}\vspace{-2.5mm}
\caption{\textbf{3D track conditioner.}
Given $\ntrack$ track pairs, $(\trk_\src^\proj, \trk_\tgt^\proj)\in\RR^{2\times\tokf\times\ntrack\times 3}$, consisting of 2D screen coordinates $xy$ and normalized disparity $z$, the conditioner first uses cross-attention to adaptively sample per-track context from source video tokens $\vidtok_\src$.
A second cross-attention then splats the sampled tokens, $\trktok^\sampled_\src$, via the source and target tracks into the respective frame spaces, yielding $[\trktok_\src, \trktok_\tgt]\in\RR^{2\tokfhw\times\tokd}$ aligned with the paired video tokens $[\vidtok_\src, \vidtok_\tgt]$. 
We use different token orientations to represent the corresponding dimensions.
\vspace{-4mm}}
\label{fig:trackconditioner}
\end{figure}
\topic{3D track conditioning.} 
Many existing motion-controlled approaches~\cite{motionprompting,diffusion-as-shader,levitor,draganything} explicitly represent motion in 2D screen space before feeding it into the model, aligning it with the target video frames.
While this provides direct guidance, it relies on hand-crafted 2D-screen-space representations, which become difficult to handle for a large number of 3D tracks where occlusions frequently occur.
In contrast, recent works~\cite{recammaster,motioncanvas} directly encode motion values into features without explicit screen-space alignment.
While these models can handle the motion signals more adaptively, they may suffer from imprecise control and scale ambiguity.
To combine the advantages of both motion encoding approaches, we propose a learnable mechanism that adaptively encodes 3D tracks to 2D screen-aligned tokens.

First, we project the source and target 3D tracks, {\small$(\trk_\src, \trk_\tgt)$}, using respective camera parameters, {\small$(\cam_\src, \cam_\tgt)$}, to the 2D screen coordinate, as {\small$(\trk^\proj_\src,\trk^\proj_\tgt)$} {\small$\in\RR^{2\times\nframe\times\ntrack\times 3}$}, where the projected tracks' $z$ values are normalized to range $[0, 1]$ in disparity space.
Consequently, the projected 3D tracks represent relative camera motion through the coordinates in the screen space.
We then temporally downsample these projected 3D tracks to {\small$\RR^{2\times\tokf\times\ntrack\times 3}$} via nearest-neighbor sampling and employ our 3D track conditioner to transform the 3D track coordinates to track tokens, $[\trktok_\src, \trktok_\tgt]\in\RR^{2\tokfhw\times\tokd}$ to align with the video tokens, $[\vidtok_\src, \vidtok_\tgt]$  (Fig.~\ref{fig:trackconditioner}).

The core function of the 3D track conditioner is to use the projected tracks to sample visual context from the source video tokens $\vid_\src$, and splat the context back to the source and target frame spaces to establish source-target correspondence.
A similar approach, TrajAttn~\cite{trajattn}, employs nearest-neighbor sampling and splatting in hidden states with 2D tracks for I2V generation.
However, this direct sampling is not robust to noisy and frequently occluded 3D tracks.
Instead, our method, inspired by Tracktention~\cite{tracktention}, uses coordinate-based cross-attention for sampling and splatting. 
This provides an adaptive mechanism to encode a variable number of 3D tracks in 2D space.

For the cross attention, we first apply positional encoding to map the 3D track's $xyz$ into token embeddings,
yielding $\pe_{\{\src,\tgt\}}^{xyz}\in\mathbb{R}^{\tokf\ntrack\times\tokd}$. 
Then, to transfer visual context, we use the source coordinate embedding $\pe_\src^{xyz}$ as the query to sample per-frame source video tokens $\vidtok_\src\in\RR^{\tokfhw\times\tokd}$ for each track, followed by Transformer~\cite{transformer} blocks that aggregate temporal information within each track:
\vspace{-1mm}\begin{equation}\vspace{-1mm}
    \trktok_\src^\sampled= \mathrm{Transformer} \left(\attn\left(\pe^{xyz}_\src,\grid, \vidtok_\src\right)\right),
\end{equation}
where the attention key, {\small $\grid\in\RR^{\tokfhw\times\tokd}$}, denotes the positional encoding of $xy$-grid coordinates with $z=0$.

The sampled tokens, $\trktok_\src^\sampled\in\RR^{\tokf\ntrack\times\tokd}$,
which carry visual context from the source video, are then splatted back to both source and target video spaces using the source and target coordinate embedding $\pe_{\{\src,\tgt\}}^{xyz}$, yielding $[\trktok_\src, \trktok_\tgt]\in\RR^{2\tokfhw\times\tokd}$.
In reverse to the sampling step, the splatting of each branch is achieved by another cross-attention:
\vspace{-1.2mm}\begin{equation}\vspace{-1.2mm}
    \trktok_{\{\src,\tgt\}} = \attn \left( \grid, \pe^{xyz}_{\{\src,\tgt\}}, \trktok^\sampled_\src \right).
\end{equation}

Both cross-attention sampling and splatting adaptively retrieve values from corresponding 3D coordinates, facilitating the robustness to noisy point tracks.
Additionally, while Tracktention~\cite{tracktention} injects an attention bias from input tracks to guide the attention operations, we \emph{omit} this bias term as we found that it is sensitive to noisy tracks.

While the splatting assigns values to $xy$ locations, the depth information is not explicitly splatted, which limits 3D awareness. 
Thus, we additionally apply positional encoding to tracks' $z$ values, as {\small$\zpe^{z}_{\{\src,\tgt\}}\in\RR^{\tokf\ntrack\times\tokd}$}, which are added to the sampled tokens {\small$\trktok^\sampled_\src$} before the splatting branches.

Lastly, the obtained track token pairs, $[\trktok_\src, \trktok_\tgt] \in \RR^{2\tokfhw\times\tokd}$, are element-wise added to the video tokens $[\vidtok_\src, \vidtok_\tgt]$, then concatenated to feed to the DiT blocks for motion control.
Notably, we \emph{do not use} visibility labels from 3D track estimation, as visibility of edited tracks becomes ambiguous after 3D manipulation. 
Instead, all source tracks are used, regardless of occlusion, allowing the model to implicitly reason about visibility and occlusion.


\subsection{Two-Stage Training}
\label{sec:training}
\begin{figure}
\centering
\includegraphics[trim={0 0 0 3mm},clip,width=\linewidth]{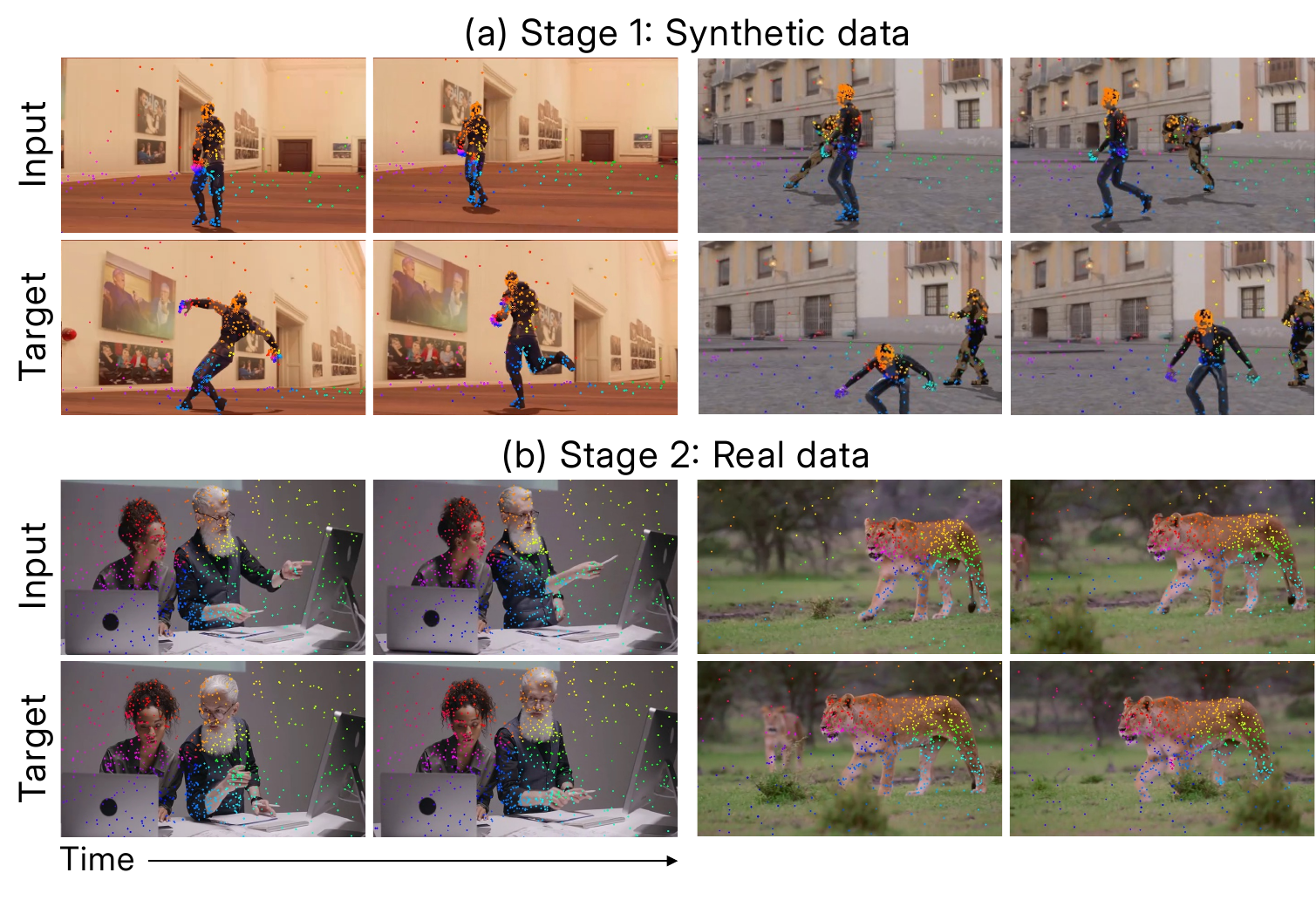}\vspace{-4mm}
\caption{\textbf{Training data.}
(a) Our model is first fine-tuned on the synthetic data with ground-truth point tracks to learn motion control. Each video pair shares the same objects and background scenes but differs in object actions and camera motions.
(b) We continue fine-tuning on real data by sampling two non-contiguous clips from a monocular video, leveraging its natural motion to scalably simulate joint camera and object motion changes.
\vspace{-3mm}}
\label{fig:traindata}
\end{figure}
\begin{figure}
\centering
\includegraphics[width=\linewidth]{figures/pdfs/apps_cam_obj_motion_edit-2cases.pdf}\vspace{-4.5mm}
\caption{\textbf{Joint camera and object motion editing.}
Our method enables the editing of camera and/or object motion using edited camera poses and 3D point tracks (visualized in corner insets).
\vspace{-5mm}}
\label{fig:apps_cam_obj_motion_edit}
\end{figure}
\begin{figure*}
\centering
\includegraphics[trim={0 0 0 3mm},clip,width=\linewidth]{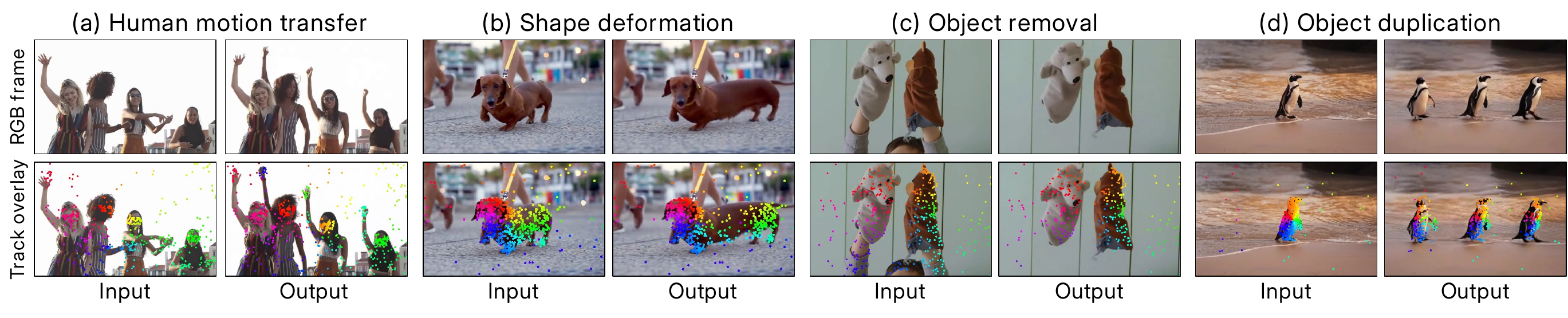}\vspace{-4mm}
\caption{{\textbf{Additional applications.}}
Our model leverages flexibly manipulated 3D tracks for diverse editing tasks.
(a) We achieve complex multi-dancer synchronization by transferring human poses via SMPL-X~\cite{smplx}.
The point tracks also enables (b) shape deformation for general objects, while (c) object removal and (d) duplication are accomplished by moving tracks off-screen or repeating them, respectively.
}\vspace{-5mm}
\label{fig:apps_others}
\end{figure*}
\begin{figure}
\centering
\includegraphics[trim={0 0 0 0},clip,width=\linewidth]{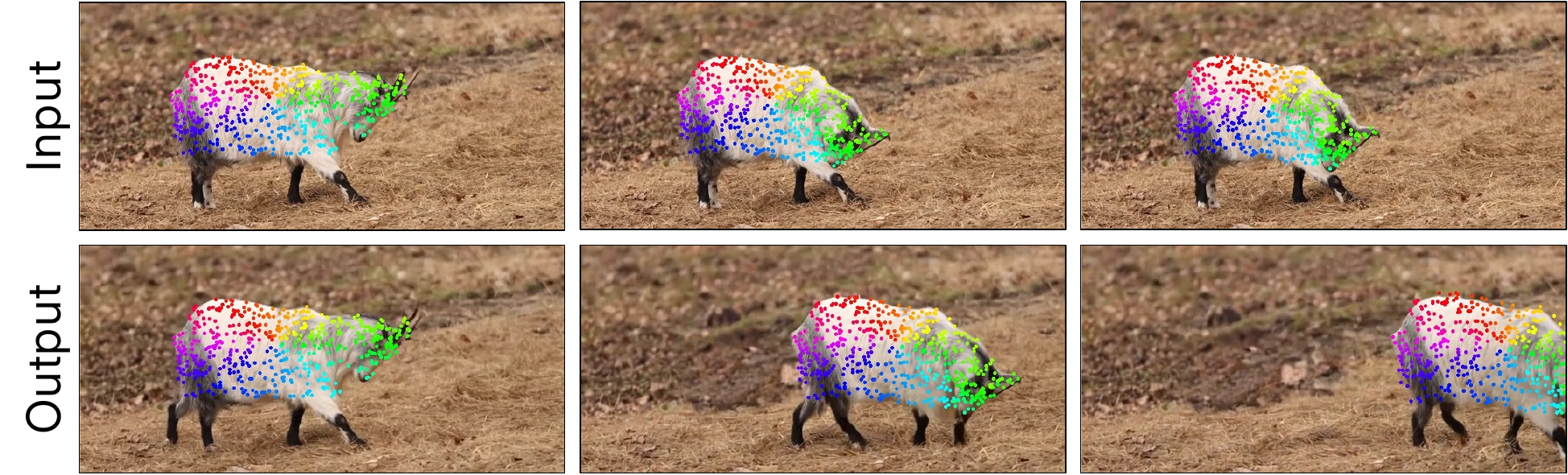}\vspace{-3mm}
\caption{{\textbf{Handling partial tracks.}}
By specifying only the body motion (moving right) via a bounding box and removing leg tracks,
our model synthesizes correct leg motion without explicit controls on the legs. Background tracks are hidden for clarity.
}
\label{fig:track_bypass}
\vspace{-5mm}
\end{figure}

Our V2V model is fine-tuned from a pretrained T2V model with an added 3D track conditioner initialized from scratch.
The ideal training data for this task would consist of large-scale video pairs depicting the same object and background context but with different camera and object motions, accompanied by precise 3D track pairs.
As such data is exceptionally rare in the real world, we employ a two-stage fine-tuning approach to address the scarcity, using both synthetic data and real monocular videos.

\topic{Stage 1: Synthetic data bootstrapping.}
To establish motion control capability, the first stage utilizes a synthetic dataset of video pairs from the same scene with varied motions and ground-truth 3D track pairs.
Scenes are generated in Blender using dynamic Mixamo human assets~\cite{mixamo} for the foreground and Kubric assets~\cite{kubric} for the background.
A base scene is initialized with randomly selected human characters and background texture, and then rendered into multiple clips using varied human animation scripts and camera trajectories.
Ground-truth 3D tracks are subsequently extracted from the mesh vertices~\cite{pointodyssey}.
During training, two clips are randomly selected from a base scene to form a training pair 
(Fig.~\ref{fig:traindata}a).

\topic{Stage 2: Real data fine-tuning.}
We bridge the synthetic-to-real domain gap by fine-tuning on real video pairs.
Given the scarcity of real-world videos with controlled variations in camera and object motion, we construct such pairs directly from monocular videos--a process that scales efficiently.
Each pair is formed by randomly sampling two \emph{non-contiguous} clips (with a temporal gap of 1-5 seconds) from a single dynamic video, allowing the clips to naturally capture changes in both camera and object motion (Fig.~\ref{fig:traindata}b).
For track correspondence, we employ 
3D tracking models~\cite{spatracker2,tapip3d} to estimate 2K 3D tracks across the full video.

We sample 24K dynamic videos from an internal stock video dataset, focusing on those in which dynamic objects are consistently tracked.
To counteract the smooth camera motion and similar view directions often present in this collection, we add a small portion of data from the DL3DV Dataset~\cite{dl3dv}, which provides static scenes with large, arbitrary camera movements. 
Additionally, a small collection of object-removal video pairs~\cite{gen-omnimatte} is integrated to specifically enhance the capability of object-level manipulation.

\topic{Training details.}
We employ LoRA (rank=64) fine-tuning for our V2V model, initializing from the pretrained Wan2.1-T2V-1.3B model~\cite{wan}. The training is conducted at a resolution of $H\times W=384 \times 672$ with a learning rate of $1\times 10^{-4}$ and a total batch size of 64 on 16 A100 GPUs. 
During training, we use a random sample of tracks, with the sample size varying between [500, 1000]. The two-stage fine-tuning process comprised 4K iterations for Stage 1 and 8K iterations for Stage 2. More details can be found in SM.


\subsection{Inference}
\label{sec:inference}
\begin{table*}
\caption{\textbf{Quantitative comparison on joint camera and object motion on DyCheck~\cite{dycheck}.}
We report full-frame and masked (covisible areas only~\cite{dycheck}) metrics, averaged across 12 scenes.
The \colorbox{scorered}{best} and \colorbox{scoreyellow}{second-best} scores are highlighted.
Some methods use ground-truth (GT) information for their inputs.
GT $1^{\mathrm{st}}$ frame denotes using the first frame of the target GT video.
Methods marked with $^{*}$ use the estimated flow to the GT video to warp the input.
$^{*}$TrajAttn~\cite{trajattn} takes warped video input using the extension of NVS-Solver~\cite{nvssolver}.
}
\vspace{-3mm}
\resizebox{
\linewidth}{!}{
\begin{tabular}{llll|ccc|ccc}
\toprule
 &  &  &  & \multicolumn{3}{c|}{Full frame} & \multicolumn{3}{c}{Masked} \\
\multirow{-2}{*}{Method} & \multirow{-2}{*}{Base model} & \multirow{-2}{*}{Type} & \multirow{-2}{*}{Privilege GT information} & PSNR $\uparrow$ & SSIM $\uparrow$ & LPIPS $\downarrow$ & mPSNR $\uparrow$ & mSSIM $\uparrow$ & mLPIPS $\downarrow$ \\ \midrule
TrajAttn~\cite{trajattn} & SVD~\cite{stablevideodiff} & I2V + track & GT 1$^{\mathrm{st}}$ frame & 12.16 & .345 & .592 & 12.25 & .697 & .416 \\
DaS~\cite{diffusion-as-shader} & CogVX~\cite{cogvideox} & I2V + track & GT 1$^{\mathrm{st}}$ frame & 13.69 & .390 & .563 & 13.75 & .724 & .391 \\
PaC~\cite{perception-as-control} & SD1.5~\cite{stablediff} & I2V + track & GT 1$^{\mathrm{st}}$ frame & 12.78 & .371 & .544 & 12.53 & .737 & .385 \\
ATI~\cite{ati} & Wan~\cite{wan} & I2V + track & GT 1$^{\mathrm{st}}$ frame & 13.67 & .371 & \cellcolor[HTML]{FFF1CD}.468 & 14.18 & .711 & \cellcolor[HTML]{FFF1CD}.312 \\ \midrule
GEN3C~\cite{gen3c} & Cosmos~\cite{cosmos} & V2V + inpaint & None & 12.85 & .398 & .567 & 13.49 & .727 & .373 \\
GEN3C~\cite{gen3c}* & Cosmos~\cite{cosmos} & V2V + inpaint & GT 1$^{\mathrm{st}}$ frame + flow to GT video & 13.61 & .406 & .517 & 13.98 & .740 & .339 \\
TrajCrafter~\cite{trajcrafter} & CogVX~\cite{cogvideox} & V2V + inpaint & None & 12.08 & .383 & .585 & 13.10 & .726 & .381 \\
TrajCrafter~\cite{trajcrafter}* & CogVX~\cite{cogvideox} & V2V + inpaint & Flow to GT video & 11.98 & .376 & .593 & 12.82 & .724 & .383 \\ \midrule
ReVideo~\cite{revideo} & SVD~\cite{stablevideodiff} & IV2V + track + inpaint & GT 1$^{\mathrm{st}}$ frame & 13.11 & .407 & .649 & 13.05 & .727 & .443 \\
ReVideo~\cite{revideo}* & SVD~\cite{stablevideodiff} & IV2V + track + inpaint & GT 1$^{\mathrm{st}}$ frame + flow to GT video & 13.18 & \cellcolor[HTML]{FFF1CD}.417 & .644 & 13.18 & .730 & .434 \\
TrajAttn~\cite{trajattn}+~\cite{nvssolver}* & SVD~\cite{stablevideodiff} & IV2V + track + inpaint & GT 1$^{\mathrm{st}}$ frame + flow to GT video & \cellcolor[HTML]{FFF1CD}13.94 & .416 & .549 & \cellcolor[HTML]{FFF1CD}14.94 & \cellcolor[HTML]{FFF1CD}.741 & .351 \\ \midrule
\NAME{} (Ours) & Wan~\cite{wan} & V2V + track & None & \cellcolor[HTML]{FFCCC9}14.80 & \cellcolor[HTML]{FFCCC9}.424 & \cellcolor[HTML]{FFCCC9}.406 & \cellcolor[HTML]{FFCCC9}15.99 & \cellcolor[HTML]{FFCCC9}.747 & \cellcolor[HTML]{FFCCC9}.247 \\ \bottomrule
\end{tabular}
}
\label{tbl:dycheck_joint}
\vspace{-5mm}
\end{table*}

We demonstrate that our \NAME{} supports a wide range of video editing tasks during inference. 
We strongly encourage readers to view our video results in our \href{https://edit-by-track.github.io}{webpage}. 

\topic{Point track preprocessing.}
For a test video, we begin by using SAM2~\cite{sam2} to obtain foreground object masks, and 3D tracking models~\cite{tapip3d,spatracker2} estimate camera parameters, depths, and 3D tracks across all frames.
We then leverage the segmentation masks to label point tracks by object for object-level track editing.
Some estimated background tracks may exhibit temporal jitter in static scenes, leading to slight distortion in certain viewpoint editing tasks (\eg, stationary views). 
Hence, we can optionally fix background points as static using their segmentation labels in this step.

\topic{Application 1: Joint camera and object motion editing.}
Since 3D point track estimation disentangles scene-object motion from camera motion in a video, users can independently edit their desired camera and/or object motion.
For example, a user can manipulate only the 3D tracks to change object movement while preserving the original camera motion, or vice versa.
Fig.~\ref{fig:apps_cam_obj_motion_edit} shows that our model effectively handles edited viewpoints and tracks, generating the desired video outputs,
even under unrealistic editing scenarios.

\topic{Application 2: Human motion transfer.}
We leverage the desirable properties of the SMPL-X $\text{\cite{smplx}}$ representation to enable human motion transfer.
First, we estimate SMLP-X parameters~\cite{prompthmr} for humans in the input video.
Motion is then transferred by swapping the local pose parameters ($\theta$) from a source human, while keeping the target's shape parameters ($\beta$) and global pose.
Finally, we reconstruct mesh vertices from the original and edited SMPL-X models to obtain corresponding point tracks.
Our model effectively handles these tracks to achieve complex human motion transfer like synchronizing multiple dancers' movements (Fig.~\ref{fig:apps_others}a), extending prior work on 2D-layer-based retiming~\cite{retiming}.

\topic{Application 3: Shape deformation.}
Our method also supports non-rigid shape deformation for general moving objects.
In practice, users select groups of points using 2D bounding boxes, transform these groups. The unselected points are then interpolated via an operation similar to Linear Blending Skinning.
Fig.~\ref{fig:apps_cam_obj_motion_edit}b shows this application with a body-shape transformation of a walking dog.

\topic{Application 4: Object removal and duplication.}
The point track inputs support additional applications, such as object removal and duplication (Fig.~\ref{fig:apps_others}c and d).
Object removal is achieved by moving the target object's points out of the frame boundaries. 
Notably, our approach removes objects while simultaneously supporting viewpoint changes, a capability not present in existing inpainting-based methods~\cite{gen-omnimatte,vace}.
For object duplication, we replicate both source and target tracks of an object to establish a new correspondence, where the new duplicate can also be manipulated by different 3D transformations.

\topic{Handling partial track input.}
While 3D tracks offer precise control, they can be unintuitive for novice users to manage multiple tracks.
Our model mitigates this by supporting partial point tracks, avoiding detailed track editing.
For example (Fig.~\ref{fig:track_bypass}), a user can use bounding boxes to select the points on a goat's body and specify a key target motion (\eg, a simple shift) while omitting the legs. The model then synthesizes plausible leg motion without manual specification.

Our model processes an 81-frame, $672\times 384$ video in ~4.5 minutes on an A100 GPU, using 50 sampling steps and a text classifier-free guidance (CFG)~\cite{cfg} scale of 5.
\section{Experiments}
\label{sec:experiments}
\begin{table}[]
\caption{\textbf{Quantitative comparison on in-the-wild videos.}
We compare with track-conditioned methods
on a test set of 100 videos randomly sampled from MiraData~\cite{miradata}.
We report PSNR, SSIM, LPIPS, FVD for visual quality, and EPE for track control. 
}
\vspace{-3mm}
\centering
\resizebox{
\linewidth}{!}{
\begin{tabular}{l|lr|ccccr}
\toprule
Method & Base model & \# params & PSNR $\uparrow$ & SSIM $\uparrow$ & LPIPS $\downarrow$ & \multicolumn{1}{l}{FVD $\downarrow$} & \multicolumn{1}{l}{EPE $\downarrow$} \\ \midrule
ReVideo~\cite{revideo} & SVD~\cite{stablevideodiff} & 1.5B & 17.10 & .599 & .388 & 438.50 & 24.00 \\
TrajAttn~\cite{trajattn} & SVD~\cite{stablevideodiff} & 1.5B & 17.59 & .617 & .359 & 396.40 & 26.54 \\
DaS~\cite{diffusion-as-shader} & CogVX~\cite{cogvideox} & 5B & 18.15 & .599 & .315 & 393.32 & 17.92 \\
PaC~\cite{perception-as-control} & SD1.5~\cite{stablediff} & 0.9B & 17.51 & .581 & .336 & 345.29 & 31.23 \\
ATI~\cite{ati} & Wan~\cite{wan} & 14B & \cellcolor[HTML]{FFF1CD}19.07 & \cellcolor[HTML]{FFF1CD}.635 & \cellcolor[HTML]{FFF1CD}.244 & \cellcolor[HTML]{FFCCC9}268.80 & \cellcolor[HTML]{FFF1CD}11.44 \\
Ours & Wan~\cite{wan} & 1.3B & \cellcolor[HTML]{FFCCC9}19.55 & \cellcolor[HTML]{FFCCC9}.657 & \cellcolor[HTML]{FFCCC9}.236 & \cellcolor[HTML]{FFF1CD}306.44 & \cellcolor[HTML]{FFCCC9}6.12 \\ \bottomrule
\end{tabular}
}
\vspace{-5mm}
\label{tbl:mira}
\end{table}

\topic{Baseline.}
We conduct extensive comparisons with existing motion-controlled video diffusion methods, categorized into three groups, as no prior V2V method directly supports joint camera and object motion editing.
(i) Track-conditioned I2V methods~\cite{ati,perception-as-control,diffusion-as-shader,trajattn} generate videos from a single image and motion tracks. 
We input the ground-truth first frame instead of the original video. 
(ii) Camera-controlled V2V methods~\cite{trajcrafter,gen3c} are designed solely for viewpoint manipulation by inpainting a warped target view. 
To test their limits, we supply them with input videos warped by joint camera and object motion.
(iii) ReVideo~\cite{revideo} and TrajAttn~\cite{trajattn} with an NVS-Solver~\cite{nvssolver} extension perform motion editing by inpainting a masked video, guided by a first frame and motion tracks.
The baselines~\cite{ati,perception-as-control,revideo} use few point tracks; thus, we adopt their respective methods to select a subset from our input tracks.

\topic{Evaluation method.}
Due to the lack of ground-truth real video pairs for joint camera and object motion manipulation, we simulate the in/output pair by selecting two \emph{non-contiguous} clips from a monocular video to define the target motion change.
We use widely adopted metrics, PSNR, SSIM~\cite{ssim}, and LPIPS~\cite{lpips} to measure the difference between the (pseudo-)ground-truth and generated videos.

\subsection{Evaluation on the DyCheck iPhone Dataset}
\label{sec:dycheck}
DyCheck~\cite{dycheck} provides synchronized multi-view videos with depth and camera-pose annotations. 
These sequences are captured in dynamic scenes using one moving camera and two stationary cameras, enabling us to validate manipulations of camera and object motion, either jointly or independently.
Our primary task, joint camera and object motion control, is tested by extracting two clips from the moving-camera video, which naturally contains both motion types  (12 scenes). 
The evaluation of disjoint motion control is provided in SM.
We report PSNR, SSIM, LPIPS~\cite{lpips}, and the respective masked scores~\cite{dycheck}, which evaluate only the co-visible areas between input and output.

Table~\ref{tbl:dycheck_joint} shows the quantitative results for the joint motion manipulation task.
Track-conditioned I2V methods~\cite{ati,perception-as-control,diffusion-as-shader,trajattn}, despite using the ground-truth first frame for motion animation, lack the full scene context from the input video, leading to notable hallucinations and object distortions.
Inpainting-based camera-controlled V2V methods~\cite{gen3c,trajcrafter} struggled to synthesize the target object motion when given the original point-cloud-warped video.
While we attempted to improve their performance by providing a warped video derived from estimated optical flow~\cite{raft} between the ground-truth and input frames, large viewpoint changes often led to flow estimation failures, preventing overall score improvement.
Finally, although ReVideo~\cite{revideo} and extended TrajAttn~\cite{trajattn} use the ground-truth first frame and partial background video, they lack full spatiotemporal context of the input video.
In contrast, our method outperforms prior approaches on all metrics by jointly managing object and viewpoint changes while preserving original context (See SM for visual comparisons).

\subsection{Quantitative Evaluation on In-the-Wild Videos}
\begin{figure}
\centering
\includegraphics[width=\linewidth]{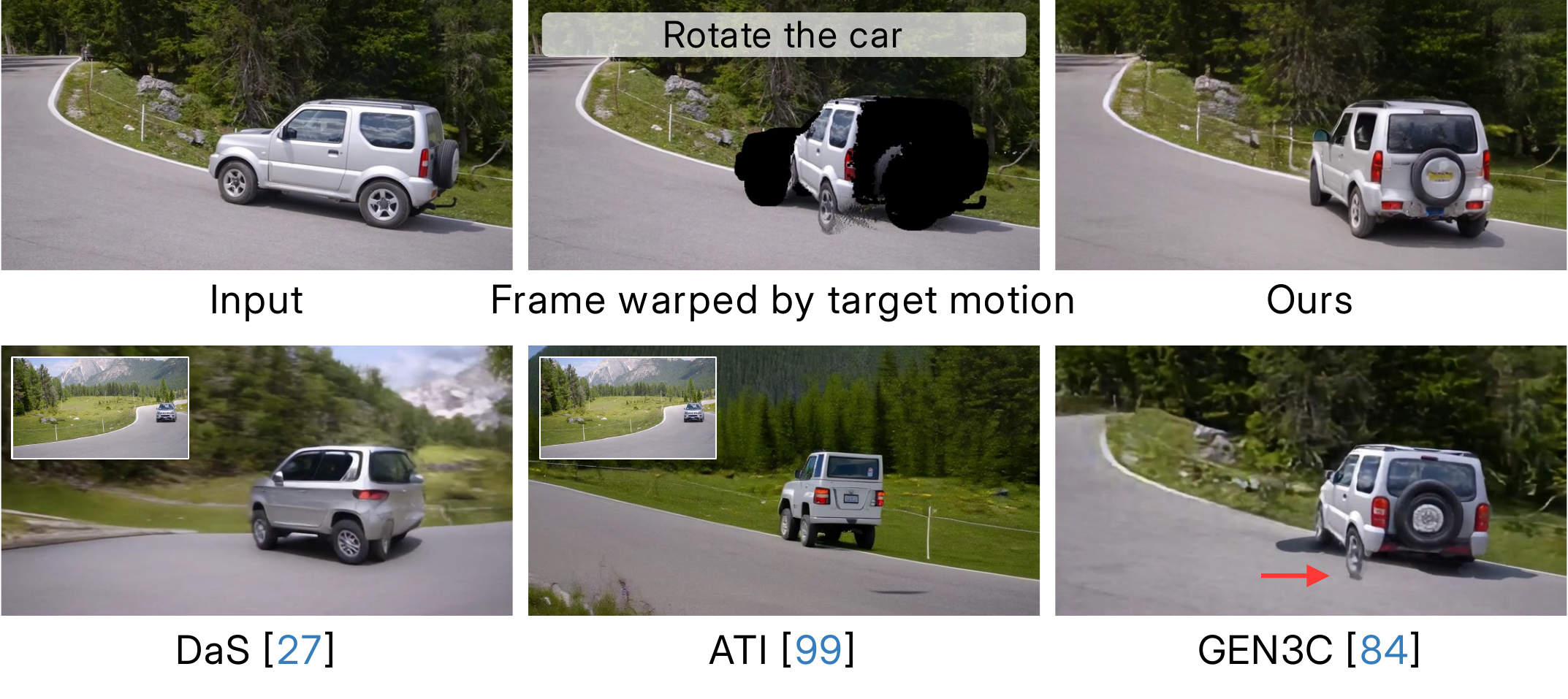}\vspace{-1.5mm}

\vspace{-2.5mm}
\caption{\textbf{Visual comparisons on video editing.}
We edit a DAVIS~\cite{davis} video with a 3D object rotation, using the target motion-warped as reference.
I2V methods~\cite{ati,diffusion-as-shader} lose context outside of the input first frame (corner insets).
GEN3C~\cite{gen3c} inputs the warped video but fails to correct the shadow of the edited object (red arrow).
See SM for additional in-the-wild results.
\vspace{-4mm}}
\label{fig:comparison_editing}
\end{figure}

We compare against track-conditioned methods~\cite{diffusion-as-shader,ati,perception-as-control,revideo,trajattn} on a test set of 100 monocular videos randomly sampled from MiraData~\cite{miradata} to form in/output pairs with joint motion changes.
In Table~\ref{tbl:mira}, we also evaluate overall visual quality using the Fréchet Video Distance (FVD)~\cite{fvd} and motion control using the End-Point Error (EPE), which measures the $L_2$ distance between estimated 2D tracks~\cite{cotracker3} in the generated versus ground-truth video. 
ATI~\cite{ati} achieves a low FVD score by using the ground-truth first frame and a massive 14B base model. However, it fails to preserve the full context of the input video or adhere to the input tracks.
In contrast, our 1.3B model achieves superior overall results and the best track control by better leveraging spatiotemporal information from the input.

\subsection{Comparisons on Video Editing}
We compare our method against baseline approaches on 10 real-world editing examples.
These edits involve modifying camera and/or object motion, with object manipulation performed on 3D tracks and depth-unprojected point clouds.
The edited point cloud, reprojected to target views, provides the warped input for GEN3C~\cite{gen3c}.
Additionally, we ensure that the first frame remains unedited to allow I2V methods~\cite{diffusion-as-shader,ati} to generate target motion from the first frame--a constraint our method does not assume.
As shown in Fig.~\ref{fig:comparison_editing}, I2V methods~\cite{ati,diffusion-as-shader} suffer from severe visual context loss, and GEN3C~\cite{gen3c} struggles with object editing, leaving noticeable shadows at the original location.
Full video results and quantitative human evaluation are provided in SM.

\subsection{Ablation Study}
\label{sec:ablation}
\topic{3D track conditioning.}
In Table~\ref{tbl:abla_trackenc}, we first compare our adaptive cross-attention sampling and splatting with a na\"ive baseline that employs a fixed Gaussian kernel (used as a bias term in Tracktention~\cite{tracktention}) for direct sampling and splatting, making it more sensitive to the noisy tracks than our adaptive approach.
Moreover, 3D track inputs offer crucial depth cues that 2D inputs lack, improving performance on large viewpoint changes on DyCheck~\cite{dycheck}.
Finally, injecting tracks' $z$ embeddings $\zpe^{z}$ into sampled tokens 
enhances 3D-aware control, reducing the track control error (EPE).

\topic{Training scheme.}
Table~\ref{tbl:abla_trainscheme} validates our two-stage strategy using both synthetic and real data.
We find that training solely on synthetic data results in weak generalization, whereas training only on real data leads to poor track control.
Lastly, our two-stage strategy, which first trains on synthetic data, then fine-tunes on real data, outperforms single-stage training on a mixed synthetic-real dataset.

\begin{table}
\caption{\textbf{Ablation study on 3D track conditioning.}
Our method (bottom) adaptively handles 3D tracks using cross-attentional sampling/splatting and injects depth embeddings for 3D-aware control. Ablation configurations are detailed in Sec.~\ref{sec:ablation}.
}\vspace{-3mm}
\resizebox{
\linewidth}{!}{
\begin{tabular}{lcl|ccc|ccccr}
\toprule
Sampling & 2D/3D & \multicolumn{1}{c|}{Inject} & \multicolumn{3}{c|}{DyCheck} & \multicolumn{5}{c}{In-the-wild videos} \\
method & tracks & \multicolumn{1}{c|}{depth} & PSNR $\uparrow$ & SSIM $\uparrow$ & LPIPS $\downarrow$ & PSNR $\uparrow$ & SSIM $\uparrow$ & LPIPS $\downarrow$ & \multicolumn{1}{l}{FVD $\downarrow$} & \multicolumn{1}{l}{EPE $\downarrow$} \\ \midrule
Na\"ive & 2D &  & 13.42 & .377 & .489 & 16.69 & .569 & .335 & 413.84 & 16.18 \\
Cross-attn & 2D &  & 13.88 & .407 & .415 & 18.88 & \cellcolor[HTML]{FFF1CD}.650 & .249 & \cellcolor[HTML]{FFCCC9}305.93 & \cellcolor[HTML]{FFF1CD}7.03 \\
Cross-attn & 3D &  & \cellcolor[HTML]{FFCCC9}14.82 & \cellcolor[HTML]{FFF1CD}.423 & \cellcolor[HTML]{FFCCC9}.395 & \cellcolor[HTML]{FFF1CD}19.07 & .647 & \cellcolor[HTML]{FFF1CD}.246 & 335.85 & 7.44 \\ \midrule
Cross-attn & 3D & \multicolumn{1}{c|}{\checkmark} & \cellcolor[HTML]{FFF1CD}14.80 & \cellcolor[HTML]{FFCCC9}.424 & \cellcolor[HTML]{FFF1CD}.406 & \cellcolor[HTML]{FFCCC9}19.55 & \cellcolor[HTML]{FFCCC9}.657 & \cellcolor[HTML]{FFCCC9}.236 & \cellcolor[HTML]{FFF1CD}306.44 & \cellcolor[HTML]{FFCCC9}6.12 \\ \bottomrule
\end{tabular}
}\vspace{-2mm}
\label{tbl:abla_trackenc}
\end{table}
\begin{table}
\caption{\textbf{Ablation study on training scheme.}
Our method (bottom) first learns track control on synthetic data, then fine-tunes on real data for generalizability, outperforming all ablated settings.
}\vspace{-3mm}
\resizebox{
\linewidth}{!}{
\begin{tabular}{lll|ccc|ccccr}
\toprule
\multicolumn{1}{c}{Syn} & \multicolumn{1}{c}{Real} & \multicolumn{1}{c|}{Two} & \multicolumn{3}{c|}{DyCheck} & \multicolumn{5}{c}{In-the-wild videos} \\
\multicolumn{1}{c}{data} & \multicolumn{1}{c}{data} & \multicolumn{1}{c|}{stage} & PSNR $\uparrow$ & SSIM $\uparrow$ & LPIPS $\downarrow$ & PSNR $\uparrow$ & SSIM $\uparrow$ & LPIPS $\downarrow$ & \multicolumn{1}{l}{FVD $\downarrow$} & \multicolumn{1}{l}{EPE $\downarrow$} \\ \midrule
\checkmark &  &  & 9.61 & .286 & .706 & 11.92 & .431 & .562 & 623.32 & 24.64 \\
 & \checkmark &  & 10.62 & .311 & .669 & 12.68 & .462 & .568 & 594.35 & 63.98 \\
\checkmark & \checkmark &  & \cellcolor[HTML]{FFF1CD}13.34 & \cellcolor[HTML]{FFF1CD}.370 & \cellcolor[HTML]{FFF1CD}.483 & \cellcolor[HTML]{FFF1CD}18.75 & \cellcolor[HTML]{FFF1CD}.636 & \cellcolor[HTML]{FFF1CD}.258 & \cellcolor[HTML]{FFF1CD}347.95 & \cellcolor[HTML]{FFF1CD}6.93 \\ \midrule
\checkmark & \checkmark & \checkmark & \cellcolor[HTML]{FFCCC9}14.80 & \cellcolor[HTML]{FFCCC9}.424 & \cellcolor[HTML]{FFCCC9}.406 & \cellcolor[HTML]{FFCCC9}19.55 & \cellcolor[HTML]{FFCCC9}.657 & \cellcolor[HTML]{FFCCC9}.236 & \cellcolor[HTML]{FFCCC9}306.44 & \cellcolor[HTML]{FFCCC9}6.12 \\ \bottomrule
\end{tabular}
}\vspace{-5mm}
\label{tbl:abla_trainscheme}
\end{table}

\section{Discussion and Conclusion}
\label{sec:conclusion}

We proposed a novel video editing framework that enables joint manipulation of 3D viewpoints, object motion, and shape deformation, tasks challenging for existing methods. 
Despite its effectiveness, our approach still has a few limitations. For example, it may struggle when point tracks are densely clustered, especially for small objects, hindering accurate visual context extraction and motion control.
It may also fail to synthesize complex physical phenomena arising from the edited motions (see SM for failure cases), which reflects the limited physical grounding of current generative priors. 
We believe these limitations can be alleviated by advances in physically grounded generative models and data scaling, 
which our scalable fine-tuning on real monocular videos is positioned to leverage.
Our method enables versatile controllable video motion editing, bridging the gap between user intent and complex video synthesis.

\vspace{3mm}\noindent\textbf{Acknowledgements.} We are grateful for the valuable feedback and insightful discussions provided by Yihong Sun, Linyi Jin, Yiran Xu, Quynh Phung, Dekel Galor, Chun-Hao Paul Huang, Tianyu (Steve) Wang,  Ilya Chugunov, Jiawen Chen, Marc Levoy, Wei-Chiu Ma, Ting-Hsuan Liao, Hadi Alzayer, Yi-Ting Chen, Vinayak Gupta, Yu-Hsiang Huang, and Shu-Jung Han.


{
    \small
    \bibliographystyle{ieeenat_fullname}
    \bibliography{main}
}

\clearpage
\setcounter{page}{1}

\appendix
\noindent\textbf{\Large Appendix}\vspace{0.3cm}

We elaborate our method details (Sec.~\ref{supp:sec:method}), model analysis (Sec.~\ref{supp:sec:model-analysis}), and additional baseline comparisons (Sec.~\ref{supp:sec:comparisons}).
We also strongly encourage readers to view our \href{https://edit-by-track.github.io}{webpage} for full video results.

\tableofcontents


\section{Method Details}
\label{supp:sec:method}

\subsection{Model and Training Details}
We fine-tune our track-conditioned V2V model by using LoRA on the pre-trained DiT blocks and the 3D track-conditioner, which is initialized from scratch.
The LoRA is applied to the MLPs in all attention modules of the DiT blocks, including the projection heads for query, key, value, and the feed-forward layers.

For our 3D track conditioner, we use a cross-attention module for sampling, followed by two self-attention Transformer blocks to aggregate temporal information within each track.
Finally, the paired splatting branches use a shared-weight cross-attention model to project the processed tokens back to corresponding frame spaces.
For both sampling and splatting cross-attention modules, we feed the positional embedding of track coordinates to both MLPs of query and key.
The attention value is directly applied by the attention weight from the softmax operation without passing through an MLP beforehand.
We use sinusoidal positional encoding to map each track's four input values ($xyz$ coordinates and an existence label $\in \{0, 1\}$, see Sec.~\ref{sec:existencelabel}) to a 128-dimensional embedding, matching the dimension of a single attention head (12 heads total).

The two-stage fine-tuning takes a total of 60 hours on 16 A100-80GB GPUs with a total batch size of 64. The first stage, using synthetic data, takes 4,000 iterations ($\sim$20 hours), and the second stage, using real data, takes 8,000 iterations ($\sim$40 hours).
Our 3D track conditioner comprises 45M parameters, and the LoRA adapters for the DiT blocks account for a total of 87M parameters.


\subsection{Synthetic Data Generation}
To train our model to learn motion control, we generated a synthetic dataset of video pairs. The core principle of this dataset is that each pair shares the same objects and background scene but features different object actions and camera movements.
This design allows the model to isolate motion changes from appearance.
For the background scenes, we create backgrounds by applying textures from the Kubric~\cite{kubric} dataset onto a large, dome-shaped mesh that envelops the scene.
For the foreground dynamic objects, we exploit the Mixamo human animation asset library~\cite{mixamo}. We gathered 25 unique human characters and 64 distinct action animations, enabling a large combinatorial space of motions and appearances.

Our generation process starts by defining a base scene, which consists of a randomly selected background texture and a random number $[1, 4]$ of chosen human characters.
For each base scene, we then generate four distinct video clips, each rendered with a different combination of camera movements and human actions.
The ground-truth 3D point tracks are then extracted from the mesh vertices~\cite{pointodyssey} of the rendered scenes.
Notably, we randomly designate some foreground objects as ``background dynamics.'' These specific objects perform the same action across all four clips, and we intentionally do not extract their point tracks.
This strategy aims to train the model to preserve the motion of objects when tracks are not specified.
We generated a total of 500 base scenes, resulting in a synthetic dataset of 2,000 video clips (four clips per scene).
The training process randomly selects two clips from a base scene as the training input and ground truth each time.


\subsection{Real Data Curation}
Our training dataset for Stage 2 fine-tuning is built from several sources.
The primary component consists of 24K real monocular videos curated from a large internal video dataset.
To obtain clean data suitable for training, we first apply a filtering process to remove videos that contain dynamic, cluttered objects, such as crowds on streets, flocks of birds, and fast-moving traffic.
These 24K selected videos are then preprocessed to estimate camera pose and 3D tracks, as detailed in Sec.~\ref{sec:3dpreprocess}.

We observed that while this internal dataset provides diverse motion of scene dynamics, its camera motion is often smooth and linear. This property limits the model's ability to learn from large and arbitrary viewpoint changes.
To complement these videos and address this limitation, we augment the training set with the DL3DV~\cite{dl3dv} dataset, which specifically contains static-scene videos with large viewpoint changes.

Lastly, we integrate the object-effect-removal training pairs from Gen-Omnimatte~\cite{gen-omnimatte} (comprising a total of 46 real-video pairs) into our training set.
This enhances the model's ability to remove objects while preserving the remaining scene.
During training, we set the sampling ratios for the final data mixture to 85\% for our dynamic monocular videos, 10\% for DL3DV~\cite{dl3dv}, and 5\% for Gen-Omnimatte~\cite{gen-omnimatte}.


\subsection{Track Perturbation and Data Augmentation}
\label{sec:augmentation}
For Stage 2 fine-tuning on real monocular videos, we apply data augmentations to the video frames and 3D tracks to better match the testing distribution.
A primary challenge is that the source tracks, $\trk_\src$, estimated from monocular video are often inaccurate and noisy, particularly in their depth component.
These inaccuracies are then propagated and frequently amplified when edited into target tracks, $\trk_\tgt$.
For example, when applying a 3D rotation, a point in $\trk_\src$ with an incorrect depth value will be transformed to an incorrect 3D position, resulting in significant visual artifacts.
To address this, we apply several point track perturbations directly to the target tracks ($\trk_\tgt$) during fine-tuning to improve the model's robustness to the inevitable noise during 3D track editing.
\begin{itemize}
\item Perturbing along epipolar line: To simulate noise from inaccurate depth, we perturb up to 10\% of the target tracks.
We use the estimated camera poses $(\cam_\src, \cam_\tgt)$ to transform the target 3D tracks into the source camera's coordinate system. We then apply random jitter along the viewing ray, which ensures the same 2D projection but perturbs the depth component.
Finally, these jittered target 3D tracks are reprojected back into the target video frames, which simulates 2D misalignments along the epipolar line caused by depth errors.
\item Perturbing by random homography: Since jittering along epipolar lines can be limited in videos with small viewpoint changes, we apply an additional random homography perturbation.
We begin by sampling a subset of 3D tracks (up to 10\%) to be perturbed.
We then randomly designate one frame to serve as an anchor. Following this, a series of per-frame homography matrices is obtained using randomly chosen four tracks within the subset, each matrix defining a random transformation for its corresponding frame relative to the anchor frame.
These per-frame homographies are then applied to all tracks in each frame of the selected subset to simulate more complex and noisy tracks.
\item Linear motion: To further simulate motion noise, we select up to 10\% of the target tracks and apply a linear motion drift. This involves adding a consistent 2D velocity vector to each selected track in the 2D frame space, causing it to move uniformly in a specific direction throughout the video.
\end{itemize}
These track perturbations enhance the model's robustness to noisy 3D tracks during editing, and we provide a detailed analysis of this effectiveness in Sec.~\ref{sec:analyze_noisiness}.

In addition to 3D track perturbation, we also perform data augmentation on the two non-contiguous clips sampled from a monocular video.
\begin{itemize}
\item Source frame dropout: With simple camera motion in a training pair, the model may learn to over-rely on only the first and last source frames while ignoring intermediate content.
To prevent this, we randomly zero out (mask) up to 50\% of the source video frames, encouraging the model to utilize the visual context from the entire input video.
\item Small clip overlap: While most training pairs are sampled from non-contiguous clips, we ensure that 5\% of pairs have some temporal overlap (up to 50\% of frames). This small subset of overlapping clips, when combined with our target track jittering, improves the model's ability to robustly preserve input details.
\item Horizontal flipping: Since the source and target clips sampled from a single monocular video can be visually similar, we introduce further data variance by horizontally flipping the target clip and its corresponding tracks with a 50\% probability.
\end{itemize}
These data augmentations enhance the model's robustness, ensuring it generalizes from our monocular training data to more complex editing scenarios during inference.


\subsection{Existence Label for Object Removal}
\label{sec:existencelabel}
We introduce an existence label in the track inputs, designed specifically to control the object-removal task.
By default, for non-removal tasks, this label is set to 1 for all tracks.
To specify removal, we set the label to 0 for the target tracks $\trk_\tgt$ of an object to be removed.
During training, we utilize the object-removal data from Gen-Omnimatte~\cite{gen-omnimatte}, either by setting the existence label to 0 for the target tracks, or by moving the tracks off-screen in the target video.
At inference time, removing an object involves both setting the existence label of its associated tracks to 0 and moving those tracks off-screen.

We emphasize again that this existence label is separate from the visibility labels obtained from the 3D track estimation.
Our method \emph{does not} use visibility labels for occlusion handling since 3D motion editing can introduce the ambiguity of visibility. Therefore, the model is trained to handle all tracks regardless of their visibility, enabling it to reason about depth order and occlusion for target video edits.

\begin{figure}
\centering
\includegraphics[trim={5mm 0 0 0},clip,width=0.8\linewidth]{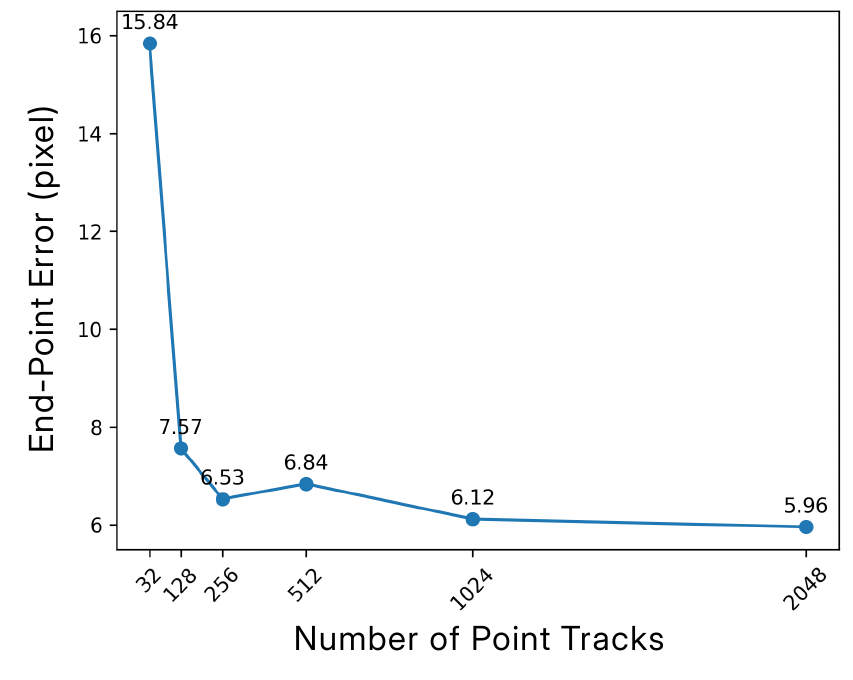}\vspace{-3mm}
\caption{\textbf{Analysis on the sparsity of track inputs.}
We evaluate End-Point Error (EPE) using 100 in-the-wild videos from MiraData~\cite{miradata} to test performance with different numbers of point tracks. All evaluations are run on the same final model, which was trained once using a random number of tracks between 500 and 1000.
}
\label{fig:sparsity}
\end{figure}
\begin{figure}
\centering
\includegraphics[width=\linewidth]{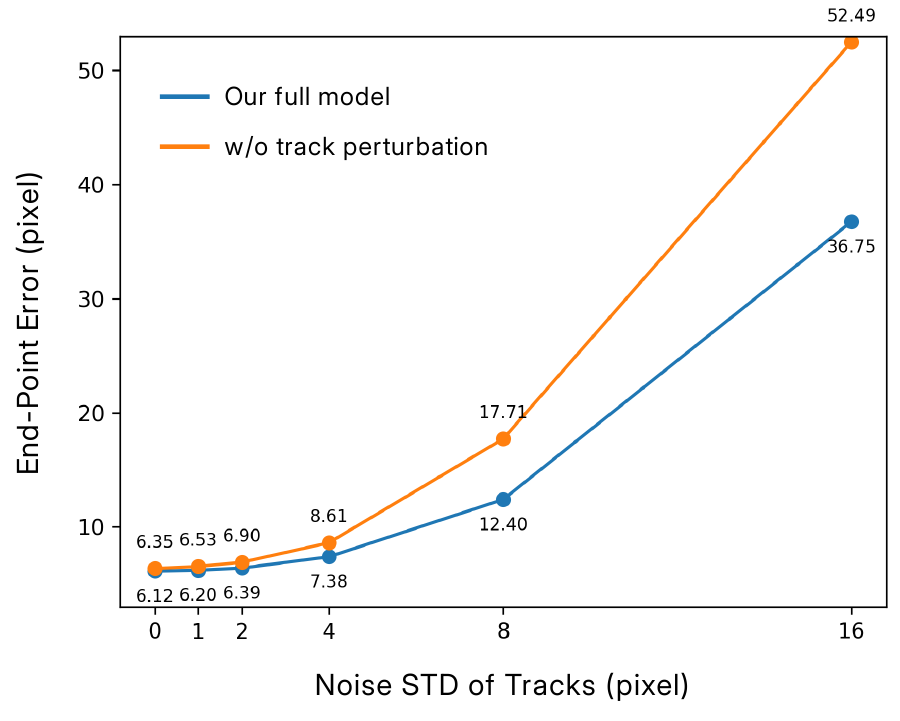}\vspace{-4mm}
\caption{\textbf{Robustness to noisy track inputs.}
Our final model, trained with track perturbation (Sec.~\ref{sec:augmentation}), is compared to an ablated model trained without it.
Both are tested with varying levels of Gaussian noise on the target tracks. Our model handles 4-pixel noise with only a 1.26-pixel increase in error, demonstrating its robustness.
}
\label{fig:noisiness}
\end{figure}

\subsection{Video Pose and 3D Point Track Estimation}
\label{sec:3dpreprocess}
We leverage the recent advancements in video-depth-pose~\cite{megasam,pi3,vipe,vggt} and 3D tracking methods~\cite{megasam} to estimate the required 3D tracks and camera parameters, including intrinsics and extrinsics, for our framework.
While some 3D methods may further apply optimization after feed-forward models to refine the 3D estimates, such as MegaSAM~\cite{megasam} for consistent depth, or SpatialTrackerV2~\cite{spatracker2} for 3D tracks, these steps can be time-consuming.
To speed up pose and track estimation, we use the fine-tuned VGGT~\cite{vggt} model in SpatialTrackerV2~\cite{spatracker2} to estimate video depth and pose, and then employ TAPIP3D~\cite{tapip3d} to obtain 3D point tracks without optimizations.
For a 400-frame training video, we sample 2000 points across 17 uniformly spaced keyframes with the same interval to ensure most video content is tracked. 
To ensure detailed foreground movements are tracked, we employ a foreground-biased sampling strategy when masks are available.
This involves densely sampling points on foreground objects while sparsely sampling the background.
The mask generation process is adapted for different phases: during training, we automatically generate masks by applying semantic segmentation to extract likely-dynamic classes; at test time, users interactively provide the foreground masks using SAM2~\cite{sam2}.
This 3D preprocessing takes approximately 4 minutes for a 400-frame video on an A100-80GB GPU.

\subsection{Editing and Inference Details}
For a given test video, we first run SAM2~\cite{sam2} to extract foreground object masks before performing video pose and 3D track estimation (Sec.~\ref{sec:3dpreprocess}). These masks serve two purposes. First, they allow us to associate point tracks with their corresponding objects (Sec.~3.3, main paper). Second, they are used to densely sample points on the foreground objects during the 3D track estimation, which is crucial for capturing detailed motions, such as those of arms and legs.

For all our demonstrated examples, the editing process exploits a Python script to apply 3D transformations to the 3D tracks and camera poses.
For more specific tasks, such as \emph{Shape deformation for body parts} or applying \emph{Partial track inputs} (\eg, removing legs' tracks to avoid detailed leg motion specification), we select a keyframe and use a 2D bounding box in the 2D frame space to select the corresponding subset of 3D points for editing.
For basic 3D track edits, we can also optionally generate a preview video by editing the depth-unprojected per-frame point clouds and warping them to the target viewpoints before running the full model.
Note that the preview video \emph{will not} be input to our model.
The 3D track and viewpoint editing process itself is very fast, typically taking less than 1 second for an 81-frame video without preview warping.
The time increases to around 30 to 60 seconds if a preview video is rendered, as that process additionally requires editing the per-frame point clouds.
In the future, we plan to develop a 3D GUI editor to make viewpoint and 3D object motion editing more accessible to general users.

Once the poses and tracks are edited, we input a random sample of 1,000 tracks to the track-conditioned V2V model. The model uses these tracks and a text prompt describing the target video to edit the video with the specified motion.
For consistency, we use a fixed random seed of 0 for all demonstrated examples and evaluations.


\section{Model Analysis}
\label{supp:sec:model-analysis}
\subsection{Robustness to Sparse Point Tracks}

Our model is trained using a random number of point tracks, ranging from 500 to 1000. We tested the trained model's robustness to track sparsity during inference by measuring the End-Point Error (EPE) on 100 in-the-wild videos from MiraData~\cite{miradata}, with results shown in Fig.~\ref{fig:sparsity}. While the model fails with extremely sparse inputs (\eg, $N=32$), due to few correspondences, it may still achieve reasonable performance with $\sim$256 tracks. Visual results are available in the \href{https://edit-by-track.github.io}{webpage}. 


\subsection{Robustness to Noisy Point Tracks}
\label{sec:analyze_noisiness}

To account for potential noise and inaccuracies in estimated 3D tracks and camera poses, we test the robustness of our model to perturbed track inputs. This experiment involves adding varying amounts of Gaussian noise to the target point tracks. We evaluate the End-Point Error (EPE) on 100 MiraData~\cite{miradata} videos, with results shown in Fig.~\ref{fig:noisiness}.
Our model, trained with track perturbation augmentation (Sec.~\ref{sec:augmentation}), is robust to handle approximately 4 pixels of noise while incurring only a 1.26-pixel increase in error.
In contrast, an ablated model trained without this augmentation is less robust, and its End-Point Error (EPE) increases more rapidly as noise levels rise.
We provide video results in the \href{https://edit-by-track.github.io}{webpage}, demonstrating the performance of our final model under various noise levels.


\subsection{Effects of Text Prompts}
\begin{figure}
\centering
\includegraphics[trim={5mm 0 0 0},clip,width=\linewidth]{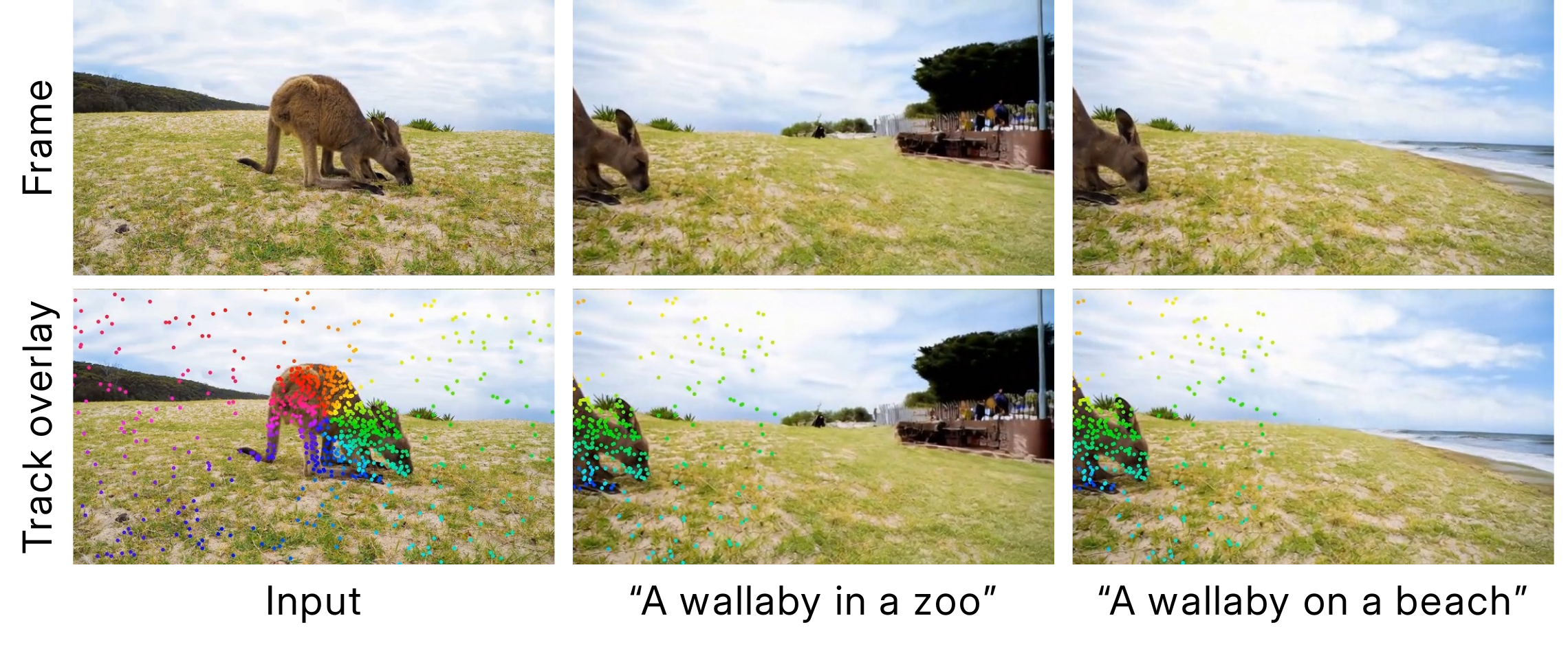}\vspace{-3mm}
\caption{\textbf{Effects of text prompts.}
The text prompt serves as a supplementary context to help generate unseen regions revealed in the novel viewpoints. As shown in the example, the right-half regions without tracks overlayed are the unseen regions.
}
\label{fig:texts}
\end{figure}
Our model utilizes 3D tracks for precise motion control, while text prompts provide supplementary context, helping to generate unseen regions from novel viewpoints  (Fig.~\ref{fig:texts}) or specific, motion-dependent effects. Please see the videos in our \href{https://edit-by-track.github.io}{project webpage}.  


\subsection{Effects of Random Seeds}
\begin{figure}
\centering
\includegraphics[trim={5mm 0 0 0},clip,width=\linewidth]{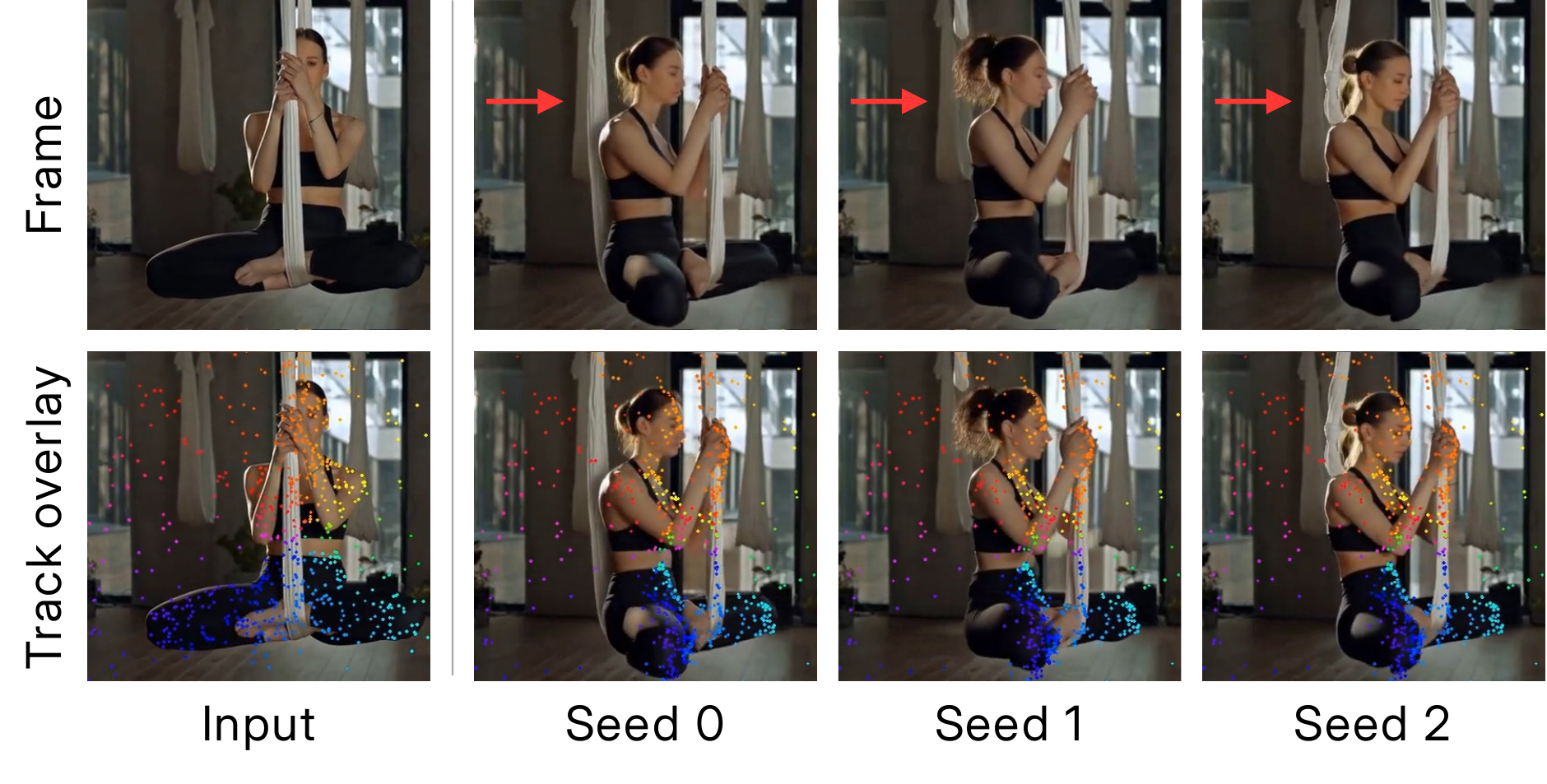}\vspace{-3mm}
\caption{\textbf{Effects of different random seeds.}
Random seeds introduce generation variations, especially in newly revealed areas (red arrows). Please note that all other examples and evaluations utilize the same fixed seed (0) for consistency.
}
\label{fig:seeds}
\end{figure}
Different random seeds produce slight variations in the generated videos, particularly in areas that are unseen from the input and revealed by the target motion (Fig.~\ref{fig:seeds}).
Note that for consistency, all our results and evaluations use a fixed seed $=0$, except for Fig.~\ref{fig:seeds}.


\subsection{Failure Cases}
\begin{figure*}
\centering
\includegraphics[width=\linewidth]{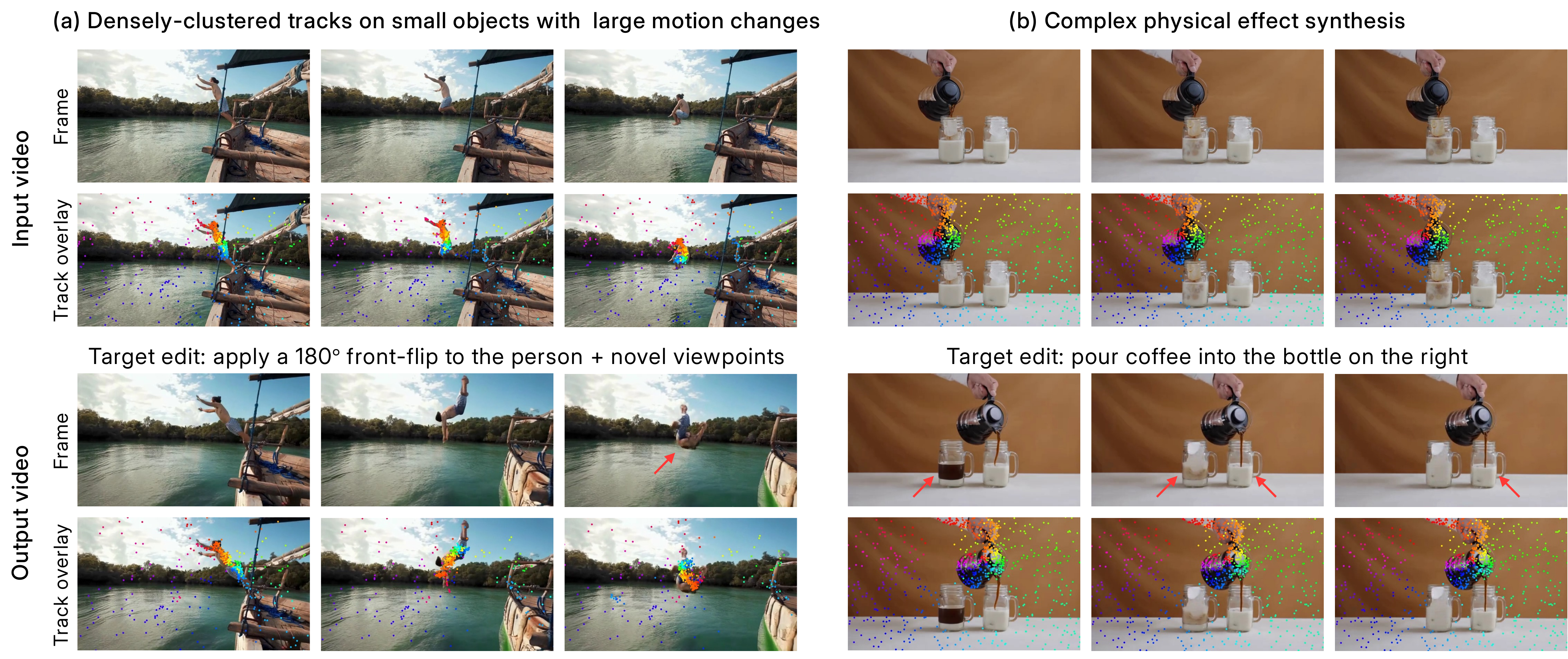}
\caption{\textbf{Limitations.}
Our model shows limitations in two main areas: (a) small objects with large motion changes (\eg, a 270\textdegree\ front-flip) can suffer from distortion when their tracks are densely-clustered and noisy; and (b) complex, motion-dependent physical effects (\eg, liquid dynamics) are not synthesized correctly, such as the failure to blend coffee and milk.
}
\label{fig:failure}
\end{figure*}
While our model demonstrates strong capability in joint camera and object motion control, it still exhibits a few limitations (Fig.~\ref{fig:failure}). First, applying large motion changes (\eg, 270$^{\circ}$ front-flipping) to small objects that are tracked by noisy, densely-clustered points can be challenging. In such cases, the model may struggle to accurately transfer visual context or precisely apply the motion condition, leading to distortion artifacts (Fig.~\ref{fig:failure}a).

Second,  although the model can synthesize plausible secondary effects associated with edited objects—such as water splashes (Fig.~2, main paper) or shadows (Fig.~9, main paper), it may fail to handle more complex physical phenomena (\eg, liquid dynamics) that arise from the target motion.
For instance, in Fig.~\ref{fig:failure}b, the model fails to synthesize the mixing of coffee and milk when the pouring action is redirected into the other milk bottle.


\section{Additional Comparisons}
\label{supp:sec:comparisons}
\begin{table}[]
\caption{\textbf{Inference runtime and resolution.}
The table reports runtime on a single A100-80GB GPU, the number of parameters in the base models, and the inference resolutions for our method and all baselines. 
}
\vspace{-3mm}
\centering
\resizebox{
\linewidth}{!}{
\begin{tabular}{l|lr|rr|r}
\toprule
\multirow{2}{*}{Method} & \multirow{2}{*}{Base model} & \multirow{2}{*}{\# params} & \multicolumn{1}{c}{Temporal} & \multicolumn{1}{c|}{Spatial resolution} & \multicolumn{1}{c}{Runtime} \\
 &  &  & \multicolumn{1}{c}{length} & \multicolumn{1}{c|}{$W\times H$} & \multicolumn{1}{c}{(min)} \\ \midrule
ReVideo~\cite{revideo} & SVD~\cite{stablevideodiff} & 1.5B & 14 & $1344\times 768$ & 1.7 \\
TrajAttn~\cite{trajattn} & SVD~\cite{stablevideodiff} & 1.5B & 25 & $1024\times 576$ & 2.0 \\
TrajAttn~\cite{trajattn}+~\cite{nvssolver} & SVD~\cite{stablevideodiff} & 1.5B & 25 & $1024\times 576$ & 6.5 \\
DaS~\cite{diffusion-as-shader} & CogVX~\cite{cogvideox} & 5B & 49 & $720\times 480$ & 4.4 \\
PaC~\cite{perception-as-control} & SD1.5~\cite{stablediff} & 0.9B & 16 & $768\times 512$ & 1.4 \\
ATI~\cite{ati} & Wan~\cite{wan} & 14B & 81 & $832\times 480$ & 14.0 \\
GEN3C~\cite{gen3c} & Cosmos~\cite{cosmos} & 7B & 121 & $1280\times 704$ & 12.7 \\
TrajCrafter~\cite{trajcrafter} & CogVX~\cite{cogvideox} & 5B & 49 & $672\times 384$ & 3.2 \\
Ours & Wan~\cite{wan} & 1.3B & 81 & $672\times 384$ & 4.5
\\ \bottomrule
\end{tabular}
}
\vspace{-5mm}
\label{tbl:runtime}
\end{table}
In this section, we present additional quantitative comparisons with existing methods. We also report the details for inference runtime and output resolutions in Table~\ref{tbl:runtime}.
We highly encourage our readers to view our \href{https://edit-by-track.github.io}{project webpage}
for full video comparisons of motion editing on in-the-wild videos.


\subsection{Evaluation on DyCheck}

DyCheck~\cite{dycheck} provides synchronized multi-view videos with depth and camera-pose annotations, captured by one moving camera and two stationary cameras.
The multi-view videos in a dynamic scene enable us to validate manipulations of camera and object motion, either jointly or independently.
We derive three distinct evaluation scenarios:
\begin{itemize}
\item \textbf{Joint camera and object motion}: evaluated by extracting two non-overlapping clips from each moving-camera video, which naturally contains both motion types (12 scenes).
\item \textbf{Camera motion only}: evaluated using a synchronized pair, where the moving-camera video serves as input and the corresponding fixed-view video as ground truth (5 scenes).
\item \textbf{Object motion only}: evaluated by extracting two non-overlapping clips from a fixed-view video (4 scenes).
\end{itemize}

\begin{figure*}
\centering
\includegraphics[width=\linewidth]{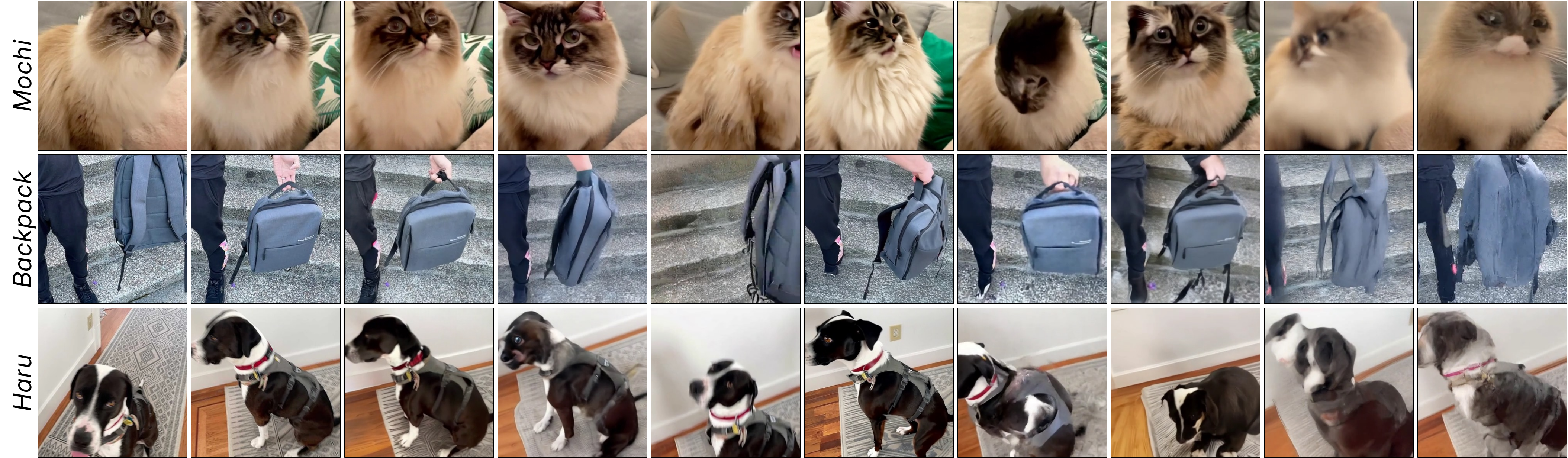}\vspace{-1.2mm}

\noindent\hspace{3mm}
\mpage{t}{0.08}{\scriptsize Input}\hfill
\mpage{t}{0.08}{\scriptsize Ground \\ truth}\hfill
\mpage{t}{0.08}{\scriptsize Ours}\hfill
\mpage{t}{0.08}{\scriptsize DaS \\ \cite{diffusion-as-shader}}\hfill
\mpage{t}{0.08}{\scriptsize PaC \\ \cite{perception-as-control}}\hfill
\mpage{t}{0.08}{\scriptsize ATI \\ \cite{ati}}\hfill
\mpage{t}{0.08}{\scriptsize GEN3C* \\ \cite{gen3c}}\hfill
\mpage{t}{0.08}{\scriptsize TrajCrafter* \\ \cite{trajcrafter}}\hfill
\mpage{t}{0.08}{\scriptsize ReVideo* \\ \cite{revideo}}\hfill
\mpage{t}{0.08}{\scriptsize TrajAttn* \\ \cite{trajattn}}\hfill

\vspace{-2.5mm}
\caption{\textbf{Visual comparison of joint motion editing on DyCheck~\cite{dycheck}.}
Our method handles large, joint camera and object motion changes, aligning with the (pseudo) ground-truth video.
Methods marked with $^{*}$ use the flow estimation between input and ground-truth videos to warp the input.
The results of TrajAttn$^{*}$ are generated with the extension of NVS-Solver~\cite{nvssolver} to take the warped video input.
}
\label{fig:comparison_dycheck}
\vspace{-2mm}
\end{figure*}
\begin{table}[]
\caption{\textbf{Quantitative comparison on camera motion only.}
We evaluate camera motion only control using the synchronized multi-view video pairs in the DyCheck Dataset~\cite{dycheck}.
}
\begin{center}
\vspace{-6mm}
\resizebox{\linewidth}{!}{
\begin{tabular}{l|ccc|ccc}
\toprule
Method & PSNR $\uparrow$ & SSIM $\uparrow$ & LPIPS $\downarrow$ & mPSNR $\uparrow$ & mSSIM $\uparrow$ & mLPIPS $\downarrow$ \\ \midrule
GEN3C~\cite{gen3c} & 13.03 & \cellcolor[HTML]{FFCCC9}.307 & \cellcolor[HTML]{FFF1CD}.544 & 14.64 & \cellcolor[HTML]{FFF1CD}.723 & .382 \\
TrajCrafter~\cite{trajcrafter} & \cellcolor[HTML]{FFF1CD}13.41 & \cellcolor[HTML]{FFF1CD}.304 & .551 & \cellcolor[HTML]{FFF1CD}14.97 & \cellcolor[HTML]{FFCCC9}.729 & \cellcolor[HTML]{FFF1CD}.358 \\
Ours & \cellcolor[HTML]{FFCCC9}13.75 & .302 & \cellcolor[HTML]{FFCCC9}.481 & \cellcolor[HTML]{FFCCC9}15.00 & .721 & \cellcolor[HTML]{FFCCC9}.316 \\ \bottomrule
\end{tabular}
}
\end{center}
\label{tbl:dycheck_camera_only}
\end{table}
\begin{table}[]
\caption{\textbf{Quantitative comparison on object motion only.} 
We compare our method with track-conditioned I2V methods on the object motion only task (with a static camera) on DyCheck~\cite{dycheck}.
Note that the baseline I2V methods directly use the ground-truth first frame as input, while our method operates without this privileged input.
}
\vspace{-6mm}
\begin{center}
\resizebox{\linewidth}{!}{
\begin{tabular}{l|ccc|ccc}
\toprule
Method & PSNR $\uparrow$ & SSIM $\uparrow$ & LPIPS $\downarrow$ & mPSNR $\uparrow$ & mSSIM $\uparrow$ & mLPIPS $\downarrow$ \\ \midrule
ReVideo~\cite{revideo} & 16.22 & .595 & .395 & 18.86 & .768 & .268 \\
TrajAttn~\cite{trajattn} & 15.01 & .415 & .370 & 17.13 & .594 & .244 \\
DaS~\cite{diffusion-as-shader} & \cellcolor[HTML]{FFF1CD}17.15 & .600 & .265 & \cellcolor[HTML]{FFF1CD}20.20 & .766 & \cellcolor[HTML]{FFF1CD}.137 \\
PaC~\cite{perception-as-control} & 16.63 & .559 & .328 & 19.42 & .729 & .218 \\
ATI~\cite{ati} & 16.79 & \cellcolor[HTML]{FFCCC9}.652 & \cellcolor[HTML]{FFF1CD}.263 & 20.11 & \cellcolor[HTML]{FFCCC9}.826 & \cellcolor[HTML]{FFCCC9}.128 \\
Ours & \cellcolor[HTML]{FFCCC9}18.11 & \cellcolor[HTML]{FFF1CD}.629 & \cellcolor[HTML]{FFCCC9}.261 & \cellcolor[HTML]{FFCCC9}21.03 & \cellcolor[HTML]{FFF1CD}.808 & .149 \\ \bottomrule
\end{tabular}
}
\end{center}
\label{tbl:dycheck_object_only}
\end{table}
\begin{figure*}
\centering
\includegraphics[width=\linewidth]{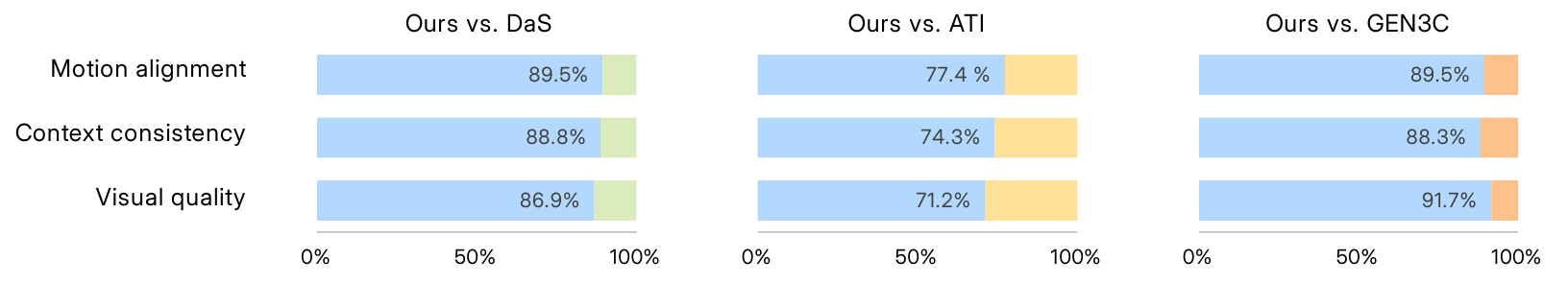}\vspace{-4mm}
\caption{\textbf{Human perceptual evaluation.}
We present the preference percentages from 42 subjects in comparison with representative methods, including track-conditioned I2V, DaS~\cite{diffusion-as-shader} and ATI~\cite{ati}, and inpainting-based V2V, GEN3C~\cite{gen3c}.
}
\label{fig:userstudy}
\end{figure*}

For the joint-motion and object-motion-only scenarios, we require training pairs consisting of two non-contiguous clips (81 frames each) from the same video. Because the provided multi-view dataset contains videos of varying durations, we select only those with sufficient length to extract such pairs. The resulting clips have an average temporal gap of 99 frames (minimum 17, median 77, maximum 222).

For fair evaluation, we first crop all videos to a landscape orientation. All methods are evaluated at a resolution of $672\times 384$.
To accommodate baselines that generate fewer than 81 frames, we apply a temporal stride (2 or 3) to the input video, ensuring access to the full video content in a single pass.
We then compute error metrics only on the corresponding subsampled output frames.
Note that ReVideo~\cite{revideo} is an exception because it outputs only 14 frames. For this method, we provide the 81-frame video subsampled with a temporal stride of 3 (yielding a 27-frame input). We then run the model twice on consecutive segments of this input and concatenate the results to obtain a 27-frame output video for evaluation.

For our primary task of joint camera and object motion control, we present the quantitative comparisons in the main paper, and the visual comparison in Fig.~\ref{fig:comparison_dycheck}.
Table~\ref{tbl:dycheck_camera_only} presents the quantitative comparisons on the camera-motion-only task. Our approach performs comparably to dedicated camera-controlled V2V methods~\cite{gen3c,trajcrafter}, which rely on dense depth warping.
Furthermore, in the object-motion-only task (Table~\ref{tbl:dycheck_object_only}), while tracked-conditioned methods~\cite{revideo,ati,perception-as-control,diffusion-as-shader,trajattn} use the ground-truth first frame, our approach achieves overall better scores due to its effective use of context from the input video.


\subsection{User Study on Video Editing}

We conduct a human perceptual evaluation with 42 subjects using the Two-Alternative Forced Choice (2AFC) method to assess 10 real-world motion editing cases, including camera and/or object motion editing.
Subjects assessed output quality based on three critical aspects: (i) alignment with desired motion, (ii) preservation of input context, and (iii) perceived visual quality.
As reported in Fig.~\ref{fig:userstudy}, our method consistently shows a higher preference over the representative baselines, DaS~\cite{diffusion-as-shader}, ATI~\cite{ati}, and GEN3C~\cite{gen3c} across all three key aspects of the video motion editing task.


\subsection{Comparison with ReCamMaster}
ReCamMaster~\cite{recammaster} is a camera-controlled video-to-video diffusion model that uses target camera extrinsics to edit viewpoints.
While it shares the same Wan2.1-T2V-1.3B base model as our method, its design is fundamentally different.
First, ReCamMaster conditions only on the target extrinsics and ignores the source camera parameters of the input video.
This design choice relies on a key assumption: the input and target output videos must share the same first frame.
In contrast, our method \emph{does not} have this same-first-frame requirement. Our training pairs are constructed from two non-contiguous clips from a monocular video, where the first frames of the clips are different by design.
Second, although directly inputting camera extrinsics is straightforward, this approach can suffer from scale ambiguity, which limits accurate camera-motion control. Our method avoids this issue by representing camera motion using screen-projected point tracks, enabling more flexible and precise control.

\end{document}